\documentclass[11pt,a4paper]{article}

\usepackage[latin1]{inputenc}
\usepackage{amsmath}
\usepackage{amsthm}
\usepackage{amsfonts}
\usepackage{amssymb}
\usepackage{authblk}
\usepackage{indentfirst}
\usepackage{url}
\usepackage{graphicx} 
\usepackage{booktabs}
\usepackage{epstopdf}
\usepackage{float}
\usepackage{caption}
\usepackage{subcaption}
\usepackage{verbatim} 
\usepackage{footnote}

\usepackage{natbib}
\usepackage[top=3cm, bottom=3cm, left=2.8cm, right=2.8cm]{geometry}
\usepackage{framed}

\usepackage{smile}

\newtheorem{remark}{Remark}
\newtheorem{assumption}{Assumption}
\newtheorem{proposition}{Proposition}
\newtheorem{lemma}{Lemma}
\newtheorem{theorem}{Theorem}

\usepackage{hyperref} 
\usepackage{booktabs}

\usepackage{algorithm}
\usepackage[algo2e]{algorithm2e}

\begin{document}

\title{The Knowledge Gradient Policy Using A Sparse Additive Belief Model}

\author{Yan Li \thanks{yanli@princeton.edu}}
\author {Han Liu \thanks {hanliu@princeton.edu}}
\author {Warren B. Powell \thanks {powell@princeton.edu}}
\affil {Department of Operations Research and Financial Engineering, Princeton University, Princeton, NJ 08544}

\date{December, 2014}
\maketitle

\begin{abstract}
We propose a sequential learning policy for noisy discrete global optimization and ranking and selection (R\&S) problems with high dimensional sparse belief functions, where there are hundreds or even thousands of features, but only a small portion of these features contain explanatory power. We aim to identify the sparsity pattern and select the best alternative before the finite budget is exhausted. We derive a knowledge gradient policy for sparse linear models (KGSpLin) with group Lasso penalty. This policy is a unique and novel hybrid of Bayesian R\&S with frequentist learning. Particularly, our method naturally combines B-spline basis expansion and generalizes to the nonparametric additive model (KGSpAM) and functional ANOVA model. Theoretically, we provide the estimation error bounds of the posterior mean estimate and the functional estimate. Controlled experiments show that the algorithm efficiently learns the correct set of nonzero parameters even when the model is imbedded with hundreds of dummy parameters. Also it outperforms the knowledge gradient for a linear model.
\end{abstract}


\section{Introduction}

The ranking and selection (R\&S) problem arises when we are trying to find the best of a set of competing alternatives through a process of sequentially testing different choices, which we have to evaluate using noisy measurements. In specific, we are maximizing an unknown function $\mu(\bx): \cX \mapsto \RR$, where $\cX \subset \RR^m$ is a finite set with $M$ alternatives. We have the ability to sequentially choose a set of measurements to estimate. Our goal is to select the best alternative when the finite budget is exhausted. We assume that experiments are time consuming and expensive. This problem arises in applications such as simulation optimization, medical diagnostics and the design of business processes. In such applications, the number of underlying parameters might be quite large; for example, we might have to choose a series of parameters to design a new material which might involve temperature, pressure, concentration and choice of component materials such as catalysts.  

The early R\&S literature assumes a lookup table belief model \citep{frazier2008knowledge,frazier2009knowledge}, but recent research has used a parametric belief model, making it possible to represent many thousands or even millions of alternatives using a low-dimensional model. Let $\bmu$ be the vector representing values of all alternatives. Linear beliefs assume the truth $\bmu$ can be represented as a linear combination of a set of parameters, that is, $\bmu=\tilde{\Xb} \balpha$, where $\balpha$ is the underlying coefficient and $\tilde{\Xb}$ represent the alternative matrix, that is, each row of $\tilde{\Xb}$ is a vector representing an alternative.

The problem is that there are many applications that is high dimensional, that is, the coefficient $\balpha$ can potentially have hundreds or even thousands of coponents. For example, in learning the accessibility profile of a large RNA molecule, the underlying weight coefficient describing the accessibility of each site is high dimensional due to the large size of RNA molecule. However, it is typically the case that only a small portion of these coefficients contain explanatory power \citep{reyes2014quantifying}.  

More generally, for these applications, we propose a sparse additive model which offers considerably more flexibility than a linear model, while recognizing that the final model will be relatively low dimensional. Sparse additive model assumes the truth takes the form
\begin{eqnarray}
\mu_i = f_1(\tilde{X}_{i1}) + f_2(\tilde{X}_{i2}) + \cdots + f_p(\tilde{X}_{ip}) + \varsigma_i, \quad \text{for } i = 1,\ldots, M, \nonumber
\end{eqnarray}
where the $f_j$s are one-dimensional smooth functions, $\varsigma_i$ is some Gaussian noise and $M$ is the number of competing alternatives. In high dimensional settings, we assume that most of the $f_j$s are zeros. If each $f_j$ is a linear function, then the sparse additive belief reduces to linear belief. In this model, we are working on a model with a potentially large number of features, most of which do not contribute significant explanatory power. Our challenge is not only to design an efficient search algorithm for identifying the best alternative, but also identify the underlying sparsity structure.

In this paper, we study high dimensional optimal learning with sparse beliefs. We first derive a knowledge gradient policy for linear models (KGSpLin) with $\ell_{1,\infty}$ group Lasso penalty. More generally, we can assume the belief function takes an additive model, which is a summation of unknown smooth functions of each feature, where only a few components are nonzero. If we approximate each smooth function by B-splines basis, the sparse additive model can be fitted using group Lasso. Therefore, KGSpLin can be naturally generalized to the knowledge gradient policy for sparse additive models (KGSpAM). Here we introduce a random indicator variable and maintain a Beta-Bernoulli conjugate prior to model our belief about which variables should be included in or dropped from the model. Additionally, in the broader class of models known as multivariate splines functional ANOVA model, tensor product B-splines can be adopted. KGSpAM can also be used in this model. 

The remainder of the paper is organized as follows. Section \ref{review} formulates the ranking and selection model in a Bayesian setting and establishes the notation used in this paper. It also highlights the knowledge gradient using both lookup table and a low dimensional linear, parametric belief model and introduces the homotopy algorithm for recursive $\ell_{1,\infty}$ group Lasso. Section \ref{KGLin} is devoted to a detailed description of the KGSpLin policy for high dimensional linear models with $\ell_{1,\infty}$ group Lasso. Section \ref{KGSpAM} generalizes the algorithm to nonparametric sparse additive belief model (KGSpAM) and also SS-ANOVA. Theoretical results are presented in Section \ref{proof}, which shows the estimation error bounds for both posterior mean and functional estimate. In Section \ref{simulation}, we test the algorithm in the context of a series of controlled experiments.

\section{Literature}
There has been a substantial literature on the general problem of finding the maximum of an unknown function where we rely on making noisy measurements to actively make searching decisions. \citet*{spall2005introduction} provides a thorough review of the literature that traces its roots to stochastic approximation methods. However, these methods require lots of measurements to find maxima precisely, which is unrealistic when measurements are very expensive.

Our problem originates from the R\&S problem, which has been considered by many authors, under four distinct mathematical formulations. We specifically consider the Bayesian formulation, for which early work dates to \citet*{raiffa1968statisticaldecision}. The other mathematical formulations are the indif-ference-zone formulation \citep{bechhofer1995design}; the optimal computing budget allocation, or OCBA \citep{chen2010stochastic,chen2012optimal}; and the large-deviations approach \citep{glynn2004large}. 

In the Bayesian formulation, this R\&S problem has received considerable attention under the umbrella of optimal learning \citep{powell2012optimal}. In this work, there are three major classes of function approximation methods: look-up tables, parametric models and nonparametric models. \citet*{gupta1996bayesian} introduces the idea of selecting an altermative based on the marginal value of information. \citet{frazier2008knowledge} extends the idea under the name knowledge gradient using a Bayesian approach which estimates the value of measuring an alternative by the predictive distributions of the means, where it was shown that the policy is myopically optimal by construction and asymptotically optimal. The knowledge gradient using a lookup table belief model approximates the function in a discrete way, without any underlying explicit structural assumption, for both uncorrelated and correlated alternatives \citep{frazier2008knowledge,frazier2009knowledge}. Another closely related idea can be found in \citet{chick2001new}, where samples are allocated to maximize an approximation to the expected value of information. \citet{negoescu2011knowledge} introduces the use of a parametric belief model, making it possible to solve problems with thousands of alternatives. For nonparametric beliefs, \citet{mes2011hierarchical} proposes a hierarchical aggregation technique using the common features shared by alternatives to learn about many alternatives from even a single measurement, while \citet*{barut2013optimal} estimates the belief function using kernel regression and aggregation of kernels. However, all the methods above assume low dimensional belief models, where the number of features is relatively small. There are applications with hundreds or even thousands of features, but where only a few features are relevant. In such settings, previous algorithms may require a lot of tedious computation on the overall features. 

Additionally, outside of the Bayesian framework, there is another line of research on sparse online learning, in which an algorithm is faced with a collection of noisy options of unknown value, and has the opportunity to engage these options sequentially. In the online learning literature, an algorithm is measured according to the cumulative value of the options engaged, while in our problem we only need to select the best one at the end of experiments. Another difference is that, rather than value, researchers often consider the regret, which is the loss of our option compared with the optimal decision in hindsight. Cumulative value/regret is appropriate in dynamic settings such as maximizing the cumulative rewards (learning while doing), while terminal value/regret fits in settings such as finding the best route in a transportation network (learn then do). Moreover, most of the algorithms in online learning are based on stochastic gradient/subgradient descent method. The key idea to induce sparsity is to introduce some regularizer in the gradient mapping \citep{duchi2009efficient,langford2009sparse,xiao2010dual,lin2011sparsity,chen2012optimal,ghadimi2012optimal}. However, a major problem with these methods is that while the intermediate solutions are sparse, the final solution may not be exactly sparse because it is usually obtained by taking the average of the intermediate solutions.

Additive models were first proposed by \citet*{friedman1981projection} as a class of nonparametric regression models and has received more attention over the decades \citep{hastie1990generalized}. In high dimensional statistics, there has been much work on estimation, prediction and model selection for penalized methods on additive model \citep{zhang2004variable,lin2006component,ravikumar2009sparse,fan2011nonparametric,guedj2013pac}. Sparsity is a feature present in a plethora of natural as well as manmade systems. In optimal learning problems, it is also natural to consider sparsity structure not only because nature itself is parsimonious but also because simple models and processing with minimal degrees of freedom are attractive from an implementation perspective. Most of the previous work on sparse additive models study it in a batch setting, but here we study it in an active learning setting, where not only observations come in recursively, but also we get to actively choose which alternative to measure.

\section{Notation and Preliminaries} \label{review}
In this section, we briefly review some results from Bayesian models for R\&S and the recursive algorithm for $\ell_{1,\infty}$ group Lasso. We start with an introduction of notation: Let $\Mb = [M_{ij}] \in \RR^{a \times d}$ and $\bv=[v_1,.\ldots,v_d]^T \in \RR^d$. We denote $\bv_I$ to be the subvector of $\bv$ whose entries are indexed by a set $I$. We also denote $\Mb_{I,J}$ to be the submatrix of $\Mb$ whose rows are indexed by $I$ and columns are indexed by $J$. For $I=J$, we simply denote it by $\Mb_{I}$ or $\Mb_{J}$. Let $\Mb_{I\ast}$ and $\Mb_{\ast J}$ be the submatrix of $\Mb$ with rows indexed by $I$, and the submatrix of $\Mb$ with columns indexed by $J$. Let $\mathrm{supp}(v):=\{j:v_j \neq 0\}$. For $0 <p <\infty$, we define the $\ell_0,\ell_p,\ell_\infty$ vector norms as
\begin{eqnarray}
\| v\|_0 := \mathrm{card}(\mathrm{supp}(v)), \| v\|_p :=(\sum_{i=1}^d | v_i|^p)^{1/p}, \mathrm{and } \| v\|_\infty := \max_{1 \leq i \leq d}|v_i | \nonumber.
\end{eqnarray}
For a matrix $\Mb$,  we define the Frobenius norm as: $\| \Mb\|_{F} : =( \sum_{i=1}^a \sum_{j=1}^d |M_{ij}|^2)^{1/2}$ and the $\ell_p$ norm to be: $\|\Mb\|_p = \max_{\|\bv\|_p = 1} \|\Mb \bv \|_p$. For any square matrix $\Mb$, let $\Lambda_{\max} (\Mb)$ and $\Lambda_{\min} (\Mb)$ be the largest and smallest eigenvalue of $\Mb$. For a summary of most symbols we use, please refer to Table \ref{table1} in Appendix A.

\subsection{The Bayesian Model for Ranking and Selection}\label{bayes}
We denote the unknown function $\mu(\bx): \cX \mapsto \RR$, where $\cX \subset \RR^m$ is a finite set with $M$ alternatives. In addition, if we have a measurement budget of $N$, our goal is to sequentially decide which alternatives to measure so that when we exhaust our budget, we have maximized our ability to find the best alternative using our estimated belief model. Here we use $\bx$ to denote the vector and $x$ to denote the corresponding alternative index, that is, $x \in \{1,\ldots,M\}$. We also use $\mu_x$ for $\mu(\bx)$. Let $\bmu=[\mu_1,\ldots,\mu_M]^T$. Under this setting, the number of alternatives $M$ can be extremely large relative to the measurement budget $N$. In a Bayesian setting, we assume $\bmu$ takes multinormal distribution
\begin{eqnarray}
\bmu \sim \cN(\btheta, \bSigma).\label{corbelief}
\end{eqnarray}
Now suppose we have a sequence of measurement decisions, $\bx^0,\bx^1,\ldots,\bx^{N-1}$ to learn about these alternatives. Here $\bx^i \in \cX$, for $i=0,\ldots,N-1$. At time $n$, if we measure alternative $x$, we observe
\begin{eqnarray}
y_x^{n+1} = \mu_x+\epsilon_x^{n+1}\nonumber,
\end{eqnarray}
where $\epsilon_x^{n+1} \sim \cN(0, \sigma^2_{x})$ and $\sigma_x$ is known. 

Initially, assume we have a multivariate normal prior distribution on $\bmu$,
\begin{eqnarray}
\bmu \sim \cN(\btheta^0,\bSigma^0)\nonumber.
\end{eqnarray}
Additionally, because decisions are made sequentially, $\bx^n$ is only allowed to depend on the outcomes of the sampling decisions $\bx^0,\bx^1,\ldots,\bx^{n-1}$. In the remainder of the paper, a random variable indexed by $n$ means it is measurable with respect to $\cF^n$, which is defined as the $\sigma$-algebra generated by $\{(\bx^0,y_{x^0}^1),(\bx^1,y_{x^1}^2),\ldots,(\bx^{n-1},y_{x^{n-1}}^n)\}$. Following this definition, we denote $\btheta^n:=\EE[\bmu|\cF^n]$, and $\bSigma^n:=\mathrm{Var}[\bmu|\cF^n]$.  It means conditionally on $\cF^n$, our posterior belief distribution on $\bmu$ is multivariate normal with mean $\btheta^n$ and covariance matrix $\bSigma^n$.  When the measurement budget of $N$ is exhausted, our goal is to find the optimal alternative, so the final decision is
\begin{eqnarray}
x^N = \argmax_{\bx \in \cX} \theta_x^N\nonumber.
\end{eqnarray}
We define $\Pi$ to be the set of all possible policies satisfying our sequential requirement; that is, $\Pi:=\{[\bx^0,\ldots,\bx^{N-1}]: \bx^n \in \cF^n \}$. Let $\EE^\pi$ indicate the expectation with respect to the prior over both the noisy outcomes and the truth $\bmu$ while the sampling policy is fixed to $\pi \in \Pi$. After exhausting the budget of $N$ measurements, we select the alternative with the highest posterior mean. Our goal is to choose a measurement policy maximizing expected reward, which can be written as
\begin{eqnarray}
\sup_{\pi \in \Pi} \EE^{\pi} \left[\max_{\bx \in \cX} \theta_x^N\right]\nonumber.
\end{eqnarray}
We work in the Bayesian setting to sequentially update the estimates of the alternatives. At time $n$, suppose we select $\bx^n=\bx$ and observe $y_{x^{n+1}}$; we can compute the $n+1$ time posterior distribution with the following Bayesian updating equations \citep{gelman2003bayesian}:
\begin{eqnarray}
\btheta^{n+1} &=& \btheta^n+\frac{y_x^{n+1}-\theta_x^n}{\sigma_x^2+\Sigma_{xx}^n} \bSigma^n \be_x,\label{varthetaupdate}\\
\bSigma^{n+1} &=& \bSigma^n - \frac{\bSigma^n \be_x \be_x^T \bSigma^n}{\sigma_x^2+\Sigma_{xx}^n}\nonumber,
\end{eqnarray}
where $\be_x$ is the standard basis vector with one indexed by $x$ and zeros elsewhere. 

\subsection{Knowledge Gradient for Linear Belief}\label{KG}
In this section, we briefly review the knowledge gradient for correlated normal beliefs (KGCB), which is a fully sequential sampling policy for learning correlated alternatives \citep{frazier2008knowledge}. Here correlated alternatives mean that the performances of different alternatives may have correlations as described in \eqref{corbelief}. We also review the knowledge gradient for a linear belief model (KGLin). It means that the belief model is linear in terms of a set of known basis functions. In this case, Bayesian updating is performed using recursive least squares \citep{frazier2009knowledge}. We represent the state of knowledge at time $n$ as: $S^n:=(\btheta^n,\bSigma^n)$. The corresponding value of being in state $S^n$ at time $n$ is
\begin{eqnarray}
V^n(S^n)=\max_{\bx' \in \cX} \theta_{x'}^n\nonumber.
\end{eqnarray}
The knowledge gradient policy is to choose the alternative that can maximize the expected incremental value,
\begin{eqnarray}
v_x^{KG,n} &=& \EE(V^{n+1}(S^{n+1}(x))-V^n(S^n)|S^n,\bx^n=\bx )\nonumber\\
&=& \EE (\max_{\bx' \in \cX} \theta_{x'}^{n+1}|S^n,\bx^n=\bx )-\max_{\bx' \in \cX} \theta_{x'}^n\nonumber
\end{eqnarray}
and
\begin{eqnarray}
\bx^{KG,n} = \argmax_{\bx \in \cX} v_x^{KG,n}.\nonumber
\end{eqnarray}
\citet{frazier2009knowledge} proposes an algorithm to compute the KG values for alternatives with correlated beliefs. First we can further rearrange equation \eqref{varthetaupdate} as the time $n$ conditional distribution of $\btheta^{n+1}$, namely, 
\begin{eqnarray}
\btheta^{n+1} = \btheta^n + \tilde{\bsigma} (\bSigma^n,\bx^n) Z^{n+1}, \label{varthetaupdateeq}
\end{eqnarray}
where
\begin{eqnarray}
 \tilde{\bsigma} (\bSigma^n,\bx)& =& \frac{\bSigma^n \be_x}{\sqrt{\sigma_x^2+\Sigma_{xx}^n}},\label{sigmatilde}\\ 
 Z^{n+1} &=& \frac{(y_x^{n+1}-\theta_x^n)}{\sqrt{\mathrm{Var}[y_x^{n+1}-\theta_x^n|\cF^n]}}\nonumber.
\end{eqnarray}
It is easy to see that $Z^{n+1}$ is standard normal when conditioned on $\cF^n$ \citep{frazier2008knowledge}. Then we substitute equation \eqref{varthetaupdateeq} into the KG formula,
\begin{eqnarray}
v_x^{KG,n} &=& \EE (\max_{\bx' \in \cX} \theta_{x'}^{n}+\tilde{\bsigma}_{x'} (\bSigma^n,\bx^n) Z^{n+1}|S^n,\bx^n=\bx )-\max_{\bx' \in \cX} \theta_{x'}^n\nonumber\\
&=& h(\btheta^n, \tilde{\bsigma} (\bSigma^n,\bx))\nonumber,
\end{eqnarray}
where $\tilde{\bsigma} (\bSigma^n,\bx) $ is a vector-valued function defined in \eqref{sigmatilde} and $\tilde{\sigma}_{x'} (\bSigma^n,\bx^n) $ indicates the component $\be_{x'}^T \tilde{\bsigma}(\bSigma^n,\bx^n)$ of the vector $\tilde{\bsigma} (\bSigma^n,\bx^n)$ and $h(\ba,\bb) = \EE[\max_i a_i + b_i Z]-\max_i a_i$ is a generic function of any vectors of the same dimension, $Z$ is a standard normal random variable. 

The expectation can be computed as the point-wise maximum of affine functions $a_i + b_i Z$ with an algorithm of complexity $O(M^2 log(M))$. It works as follows. First the algorithm sorts the sequence of pairs $(a_i, b_i)$ such that the $b_i$s are in nondecreasing order and ties in $b$ are broken by removing the pair $(a_i, b_i)$ when $b_i = b_{i+1}$ and $a_i \leq a_{i+1}$. Next, all pairs $(a_i, b_i)$ that are dominated by the other pairs, that is, $a_i + b_i Z \leq \max_{j \neq i} a_j + b_j Z$ for all values of $Z$, are removed. Thus the knowledge gradient can be computed using 
\begin{eqnarray}
v_x^{KG} = h(\ba,\bb) = \sum_{i=1,\ldots,\tilde{M}} (\tilde{b}_{i+1} - \tilde{b}_i) f\left(-\left|\frac{\tilde{a}_i-\tilde{a}_{i+1}}{\tilde{b}_{i+1} - \tilde{b}_i} \right| \right)\nonumber,
\end{eqnarray}
where $f(z) = \phi(z)+z \Phi(z)$. Here $\phi(z)$ and $\Phi(z)$ are the normal density and cumulative distribution functions respectively. $\tilde{\ba}$ and $\tilde{\bb}$ are the new vectors after sorting $a$ and $b$ and dropping off the redundant components and are of dimension $\tilde{M}$. 

If the number of alternatives is quite large, the above representation becomes clumsy. Thus if the underlying belief model has some structure, then we could take advantage of this structure to represent the model and simplify the computation. In a simple case, if $f$ has a linear form or can be written as a basis expansion, we can make it easier by maintaining a belief on the coefficients instead of the alternatives. 

\citet{negoescu2011knowledge} further extends KGCB to parametric beliefs using a linear model. Now we assume the truth $\bmu$ can be represented as a linear combination of a set of parameters, that is, $\bmu=\tilde{\Xb} \balpha$, where $\bmu \in \RR^M$ and $\balpha=[\alpha_1,\ldots,\alpha_m]^T \in \RR^m$ are random variables, $\tilde{\Xb} \in \RR^{M \times m}$ represent the alternative matrix, that is, each row of $\tilde{\Xb}$ is a vector representing an alternative. If we assume $\balpha \sim \cN(\bvartheta,\bSigma^{\bvartheta})$, this induces a normal distribution on $\bmu$ via linear transformation,
\begin{eqnarray}
\bmu \sim \cN(\tilde{\Xb} \bvartheta, \tilde{\Xb}\bSigma^{\bvartheta}\tilde{\Xb}^T)\nonumber.
\end{eqnarray} 
At time $n$, if we measure alternative $\bx^n=\bx$, we can update $\bvartheta^{n+1}$ and $\bSigma^{\bvartheta,n+1}$ recursively via Recursive Least Squares \citep{powell2012optimal},
\begin{eqnarray}
\bvartheta^{n+1} &=& \bvartheta^n +\frac{\hat{\epsilon}^{n+1}}{\gamma^n} \bSigma^{\bvartheta,n} \bx^n,\nonumber\\
\bSigma^{\bvartheta,n+1} &=& \bSigma^{\bvartheta,n} -\frac{1}{\gamma^n} (\bSigma^{\bvartheta,n} \bx^n(\bx^n)^T \bSigma^{\bvartheta,n}),\nonumber
\end{eqnarray}
where $\hat{\epsilon}^{n+1}=y^{n+1}-(\bvartheta^n)^T \bx^n$ and $\gamma^n=\sigma^2_{x}+(\bx^n)^T \bSigma^{\bvartheta,n} \bx^n$.

The linear model allows us to represent the alternatives in a compact format since the dimension of the parameters is usually much smaller than the number of the alternatives. Suppose we have tens of thousands of alternatives, without the linear model, we would need to create and update the covariance matrix $\bSigma^n$ with tens of thousands of rows and columns. With the linear model, we only need to maintain the parameter covariance matrix $\bSigma^{\bvartheta,n}$, the size of which is equal to the dimension of the parameter vector $\bvartheta$. In addition, we never need to compute the full matrix $\tilde{\Xb}\bSigma^{\bvartheta}\tilde{\Xb}^T$. We only have to compute a row of this matrix.

\subsection{A Homotopy Algorithm for Recursive $\ell_{1,\infty}$ Group Lasso}\label{relasso}
In the Bayesian updating scheme described in Section \ref{KGLin} and \ref{KGSpAM}, a recursive $\ell_{1,\infty}$ group Lasso is required, which we review in this section. When the regularization takes the $\ell_1$ norm, this regularized version with least squares loss is Lasso (least absolute shrinkage and selection operator)\citep*{tibshirani1996regression}. It is well known that Lasso leads to solutions that are sparse and therefore achieves model selection. If we consider a more general group sparsity system, which is composed of a few nonoverlapping clusters of nonzero coefficients, $\ell_{1,\infty}$ group Lasso penalty can be used to encourage correlations within groups and achieve sparsity at a group level. Here we briefly describe the recursive homotopy algorithm for $\ell_{1,\infty}$ group Lasso proposed in \citet*{chen2012recursive}. For the recursive homotopy algorithm for Lasso, one can refer to \citet*{garrigues2008homotopy}. This algorithm computes an exact update of the optimal $\ell_{1,\infty}$ penalized recursive least squares predictor. Each update minimizes a convex but nondifferentiable function optimization problem. This algorithm has been demonstrated to have lower implementation complexity than direct group Lasso solvers. It also fits the recursive setting in optimal learning. 

The $\ell_{1,\infty}$ group Lasso estimator after $n$ observations is given by
\begin{eqnarray}
\hat{\bbeta}^n =\argmin_{\bbeta \in \RR^m} \frac{1}{2} \sum_{i=1}^n  \left[(\bx^{i-1})^T \bbeta-y^i \right]^2+\lambda^n \|\bbeta \|_{1,\infty}, \label{grlasso}
\end{eqnarray}
where $(y^i, \bx^{i-1}) \in \RR \times \RR^m, i=1,\ldots,n$ are the $n$ observations. $\lambda^n$ is the regularization parameter, and
$\|\bbeta\|_{1,\infty} := \sum_{j=1}^{p} \|\bbeta_{\cG_j} \|_\infty$. $\{\cG_j\}_{j=1}^{p}$ is the group partition of the index set $\cG = \{1,\ldots,m\}$, that is, 
\begin{eqnarray}
\cup_{j=1}^{p} \cG_j = \cG, \quad \cG_j \cap \cG_j'= \emptyset \quad \mathrm{if} \quad j \neq j', \nonumber
\end{eqnarray}
and $\bbeta_{\cG_j}$ is a subvector of $\bbeta$ indexed by $\cG_j$. Let $d_j = | \cG_j|$ be the number of features in the $j$th group, and $m=\sum_{j=1}^p d_j$.  Group Lasso reduces to Lasso when each group contains only one coefficient.

At time $n$, suppose we have $\hat{\bbeta}^n$ to the Lasso with $n$ observation and we are given the next observation $(y^{n+1}, \bx^n) \in \RR \times \RR^m$. The algorithm computes the next estimate $\hat{\bbeta}^{n+1}$ via the following optimization problem. Let  $\Rb^{n-1} = \sum_{i=1}^n \bx^{i-1} (\bx^{i-1})^T$, $\br^n = \sum_{i=1}^n \bx^{i-1} y^i $. Let us define a function
\begin{eqnarray}
u(t,\lambda)=\argmin_{\bbeta \in \RR^m} \frac{1}{2} \bbeta^T( \Rb^{n-1} +t \bx^{n} (\bx^{n})^T)\bbeta -\bbeta^T( \br^{n}+t \bx^{n} y^{n+1}) + \lambda \| \bbeta\|_{1,\infty}.\nonumber
\end{eqnarray} 
We have $\hat{\bbeta}^{n}=u(0, \lambda^n)$ and $\hat{\bbeta}^{n+1}=u(1,\lambda^{n+1})$. The homotopy algorithm that computes a path from $\hat{\bbeta}^{n}$ to $\hat{\bbeta}^{n+1}$ in two steps:
\begin{itemize}
\item[1]  Fix $t=0$, vary the regularization parameter from $\lambda^n$ to $\lambda^{n+1}$ with $t=0$. This amounts to computing the regularization path between $\lambda^n$ and $\lambda^{n+1}$ using homotopy methods as the iCap algorithm done in \citet{zhao2009composite}. This solution path is piecewise linear.
\item[2]  Fix $\lambda$ and calculate the solution path between $u(0,\lambda^{n+1})$ and $u(1,\lambda^{n+1})$ using the homotopy approach. This is derived by proving that the solution path is piecewise smooth in $t$. The algorithm computes the next ``transition point" at which active groups and solution signs change, and updates the solution until $t$ reaches 1.
\end{itemize}
 
\section{Knowledge Gradient for Linear Model with $\ell_{1,\infty}$ Group Lasso}\label{KGLin}
In Section \ref{KG}, we review knowledge gradient policy for linear belief models in low dimensional settings. In this section, we derive the KG policy in a high dimensional linear model. We have $\bmu=\tilde{\Xb}\balpha$, where $\tilde{\Xb} \in \RR^{M \times m}$ is the alternative matrix and $\balpha \in \RR^m, \bmu \in \RR^M$ are random variables. Here $m$ can become relatively large and $\balpha$ is sparse in the sense that only a few components of $\balpha$ contribute to $\bmu$. However, unlike the sparsity assumption in classical frequentist statistics, we assume the sparsity structure is random; that is, the group indicator variable of which is selected or not is a random vector. Specifically, we now assume there exists some known group structure in $\balpha$, let $\bzeta=[\zeta_1,\ldots,\zeta_{p}] \in \RR^{p}$ be a group indicator random variable of $\balpha$,
\begin{eqnarray}
 \zeta_j= \left\{
  \begin{array}{l l}
    1 & \quad \text{if $\balpha_{\cG_j} \neq \textbf{0}$}\\
    0 & \quad \text{if $\balpha_{\cG_j}=\textbf{0}$}
  \end{array}, \right.
  \quad \text{for } j=1,\ldots,p.\nonumber
  \end{eqnarray}
Additionally, $\balpha$ is assumed to be sparse in the following sense,
\begin{eqnarray}
\balpha|\bzeta \sim \cN(\bvartheta, \bSigma^{\bvartheta}). \label{mixnorm}
\end{eqnarray}
Let $\cS= \{j: \zeta_j =1\}$. Thus, without loss of generality, conditioning on $\bzeta$, we can permute the elements of $\balpha$ to create the following partition,
\begin{eqnarray}
\balpha^T=[(\balpha_{\cS})^T, \textbf{0}],\nonumber
\end{eqnarray}
where $\balpha_{\cS}\sim \cN(\bvartheta_{\cS}, \bSigma^{\bvartheta}_{\cS})$. So $\bvartheta$ and $\bSigma^{\bvartheta}$ can be correspondingly partitioned
\begin{eqnarray}
\bvartheta = \begin{bmatrix}
       \bvartheta_{\cS}\\
       \textbf{0}         
     \end{bmatrix}, \quad
  \bSigma^{\bvartheta} = \begin{bmatrix}
  \bSigma^{\bvartheta}_{\cS} & \textbf{0}\\
 \textbf{0}    & \textbf{0}
  \end{bmatrix}.\nonumber
\end{eqnarray}

Here we make a critical assumption on the distribution of $\balpha$. Let us assume that conditioning on $\bzeta=\textbf{1}$, $\balpha$ has the following distribution: $\balpha|\bzeta=\textbf{1} \sim \cN(\bvartheta, \bSigma^{\bvartheta})$. Then for any other $\bzeta'$, the conditional distribution of $\balpha$ on $\bzeta'$ is normal with mean $\btheta_{\cS'}$ and variance $\bSigma^{\btheta}_{\cS'}$. Here $\cS' = \{j:\zeta'_j =1 \}$. This means that we can write all the conditional distributions of $\balpha$ through an index set $\cS$ characterized by $\bzeta$. So in the following we use both $\bzeta$ and $\cS$ as indices. We also omit the time dependent variable $n$ to simplify notations. Furthermore, as we are updating the mean and covariance matrix of a certain conditional distribution, we also update all the elements with the same index in the other distributions. That means, through all the updatings, we just need to maintain the mean and covariance matrix on $\bzeta=\textbf{1}$.

\subsection{Knowledge Gradient Policy for Sparse Linear Model}
Before deriving the sparse knowledge gradient algorithm, let us describe the Bayesian model at time $n$.  To get a Bayesian update, we can maintain Beta-Bernoulli conjugate priors on each component of $\bzeta$.  At time $n$, we have the following Bayesian model, for $j,j'=1,\ldots,p$,
\begin{eqnarray}
\balpha|\bzeta^n=\textbf{1} \sim \cN(\bvartheta^n, \bSigma^{\bvartheta,n}),\label{bayeseq1}\\
\zeta_j^n |p_j^n \sim \mathrm{Bernoulli}(p_j^n),\\
\zeta_j^n \perp \zeta_{j'}^n, \quad \mathrm{for} \quad j \neq j',\\
p_j^n|\xi_j^n,\eta_j^n \sim \mathrm{Beta}(\xi_j^n,\eta_j^n)\label{bayeseq4}.
\end{eqnarray}
At time $n$, the prior $\bzeta^n$ is a discrete random variable. Let $\bzeta^{n,1},\ldots,\bzeta^{n,N_{\bzeta}}$ be all the possible realizations of $\bzeta^n$, and $\PP(\bzeta^n=\bzeta^{n,k})=p^{n,k}, k=1,\ldots, N_{\bzeta}$. So by the Law of Total Expectation, the KG value can be computed by
\begin{eqnarray}
v_x^{KG,n} &=& \EE(V^{n+1}(S^{n+1}(x))-V^n(S^n)|S^n,\bx^n=\bx )\nonumber\\
&=& \EE (\max_{\bx' \in \cX} \theta_{x'}^{n+1}|S^n,\bx^n=\bx )-\max_{\bx' \in \cX} \theta_{x'}^n\nonumber\\
&=& \EE_{p^n} \EE_{\bzeta^n| p^n} \EE_{\balpha,\bepsilon|\bzeta^n,p^n}(\max_{\bx' \in \cX} \theta_{x'}^{n+1}|S^n,\bx^n=\bx, \bzeta^n,p^n)-\max_{\bx' \in \cX} \theta_{x'}^n\nonumber\\
&=& \sum^{N_{\bzeta}}_{k=1} {\EE_{p^n}(p^{n,k}) h(\ba^{n,k}, \bb^{n,k})}\nonumber\\
&=& \sum^{N_{\bzeta}}_{k=1} \prod_{\{j: \zeta_j^{n,k}=1\}} \frac{\xi_j^n}{\xi_j^n+\eta_j^n}\prod_{\{j: \zeta_j^{n,k}=0\}} \frac{\eta_j^n}{\xi_j^n+\eta_j^n} h(\ba^{n,k}, \bb^{n,k}),\nonumber
\end{eqnarray} 
where
\begin{eqnarray}
h(\ba,\bb) &:=& \EE[\max_i a_i + b_i Z]-\max_i a_i,\nonumber\\
\ba^{n,k} &=& \tilde{\Xb}^{n}_{\bzeta^{n,k}} \bvartheta^{n}_{\bzeta^{n,k}},\nonumber\\
\bb^{n,k} &=& \tilde{\bsigma}(\tilde{\Xb}^{n}_{\bzeta^{n,k}} \bSigma^{n,\bvartheta}_{\bzeta^{n,k}} (\tilde{\Xb}^{n}_{\bzeta^{n,k}})^T, \bx).\nonumber
\end{eqnarray}
Note that conditioning on each sample realization of $\bzeta^n$, the KG calculation is identical with KGLin.  The KG value for a sparse linear model is a weighted summation over all the possible sample realization of $\bzeta^n$, of which the weight $\EE_{p^n}(p^{n,k})$ is computed by the independent Beta distributions on all the $p_j^n$'s.  Also, if $N_{\bzeta}$ takes its largest possible value, that is $N_{\bzeta}=2^{p}$, we can re-sort the weights and approximate the knowledge gradient value by only computing ones with several top largest probabilities.

\subsection{Bayesian Update}\label{sec::update}
At time $n$ we have the Bayesian model described in \eqref{bayeseq1}-\eqref{bayeseq4}. Parallel with that, we also have the current Lasso estimate, denoted as $\hat{\bvartheta}^{n}$. The nonzero part is $\hat{\bvartheta}^{n}_{\cS}$. The covariance matrix corresponding to $\hat{\bvartheta}^{n}_{\cS}$ is denoted as $\hat{\bSigma}^{\bvartheta,n}_{\cS}$, which is Monte Carlo simulated from the first order optimality condition of the optimization problem \eqref{grlasso} (details described in Section \ref{sec::MC}). After we get the new observation, we can update to the next Lasso estimate recursively by the algorithm described in Section \ref{relasso}. Thus we have the updated Lasso estimate $\hat{\bvartheta}^{n+1}_{\cS}$ and $\hat{\bSigma}^{\bvartheta,n+1}_{\cS}$. Let $\cP^n:=\{j: \hat{\bvartheta}^n_{\cG_j} \neq 0\}$. The Bayesian updating equations are given by \citet{gelman2003bayesian}:
\begin{eqnarray}
\bSigma^{\bvartheta,n+1}_{\cS} = \left[(\bSigma^{\bvartheta,n}_{\cS})^{-1} + (\hat{\bSigma}^{\bvartheta,n+1}_{\cS})^{-1}\right]^{-1},\\\label{sigmaupdate}
\bvartheta^{n+1}_{\cS} = \bSigma^{\bvartheta,n+1}_{\cS}\left[ (\bSigma^{\bvartheta,n}_{\cS})^{-1} \bvartheta^{n}_{\cS}+(\hat{\bSigma}^{\bvartheta,n+1}_{\cS})^{-1} \hat{\bvartheta}^{n+1}_{\cS}\right],\\\label{meanupdate}
\xi_j^{n+1}=\xi_j^n+1, \eta_j^{n+1}=\eta_j^n, \quad \mathrm{for} \quad j \in \cP^{n+1},\nonumber\\
\xi_j^{n+1}=\xi_j^n, \eta_j^{n+1}=\eta_j^n+1, \quad \mathrm{for} \quad j \notin \cP^{n+1}.\nonumber
\end{eqnarray}

Now we briefly recall and summarize the random variables which play a role in the measurement process. The underlying and unknown value of alternative $x$ is denoted $\mu_x$ and parametrized by $\balpha$. Here $\balpha$ follows a ``mixture" normal distribution by \eqref{mixnorm} and $\zeta_j$ follows a conditional Bernoulli distribution with the frequency of ``in" and ``out" denoted by  $\xi_j$ and $\eta_j$. Both $\balpha$ and $\bzeta$ are randomly fixed at the beginning of the measurement process. At time $n$, $\bzeta^n$ and $\bvartheta^n$ give us the best estimate of $\balpha$. $(\bSigma^{\bvartheta,n}_{\cS})^{-1} $ is the precision with which we make this estimate. The result of our time $n$ measurement causes us to first update the Lasso solution from $\hat{\bvartheta}^n$ to $\hat{\bvartheta}^{n+1}$ and then update the mean estimate from $\bvartheta_\cS^n$ to $\bvartheta_\cS^{n+1}$, which we now know with precision $(\bSigma^{\bvartheta,n+1}_{\cS})^{-1}$. 

One may think of  $\bzeta$ and $\balpha$ as fixed and of $\bzeta^n$ as converging toward $\bzeta$ and $\bvartheta_{\cS}^n$ as converging toward $\balpha$ while some norm of the precision matrix $(\bSigma^{\bvartheta,n}_{\cS})^{-1}$ converging to infinity under some appropriate sampling strategy. It is also appropriate, however, to fix $\bzeta^n$ and $\bvartheta_{\cS}^n$ and think of $\bzeta$ and $\balpha$ as unknown quantities. Furthermore, from this perspective, the randomness of $\bzeta$ and $\balpha$ does not imply they must be chosen from Bernoulli and mixture normal distribution respectively, but instead it only quantifies our uncertain knowledge of  $\bzeta$ and $\balpha$ adopted when they were first chosen.

\subsection{Knowledge Gradient with Recursive $\ell_{1,\infty}$ Group Lasso}\label{sec::MC}
\begin{figure}[!htb]
\centering
\includegraphics[scale=0.5]{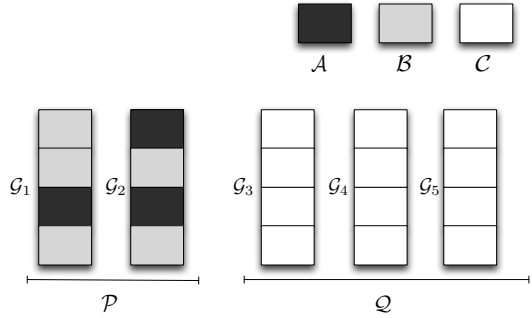}
\caption{Illustration of the partitioning of a 20 element coefficient vector $\bbeta$ into five groups of four indices. The sets $\cP$ and $\cQ$ contains the active groups and the inactive groups, respectively. Within each of the two active groups the coefficients with maximal absolute values are denoted by the black color.}
\label{omnifig:1}
\end{figure}

In this section, we first provide a technique to approximately sample the covariance matrix $\hat{\bSigma}_{\cS}^{\bvartheta,n+1}$ from the first order optimality condition in problem \eqref{grlasso}. Then we outline the knowledge gradient policy for sparse linear models in Algorithm \ref{Algorithm1},

We begin with a series of set definitions.  Figure \ref{omnifig:1} provides an illustrative example. Let us divide the entire group index into $\cP$ and $\cQ$ respectively, where $\cP$ contains active groups and $\cQ$ is the complement. For each active group $j \in \cP$, we partition the group into two parts: $\cA_j$ with maximum absolute values and $\cB_j$ with the rest of the values. That is
\begin{eqnarray}
\cA_j = \argmax_{k \in \cG_j} |\beta_k|, \quad \cB_j = \cG_j - \cA_j, \quad j \in \cP.\nonumber
\end{eqnarray}

The set $\cA$ and $\cB$ are defined as the union of the $\cA_j$ and $\cB_j$ sets, respectively,
\begin{eqnarray}
\cA = \cup_{j \in \cP} \cA_j, \quad  \cB = \cup_{j \in \cP} \cB_j.\nonumber
\end{eqnarray}
Finally, we define
\begin{eqnarray}
\cC = \cup_{j \in \cQ} \cG_j, \quad \cC_j = \cG_j \cap \cC.\nonumber
\end{eqnarray}
The $\ell_{1,\infty}$ group Lasso problem \eqref{grlasso} can also be written as
\begin{eqnarray}
\bbeta^n = \argmin_{\bbeta \in \RR^m} \frac{1}{2} \bbeta^T \Rb^{n-1} \bbeta - \bbeta^T \br^n + \lambda^n \|\bbeta \|_{1,\infty}. \label{grlasso2}
\end{eqnarray}
This optimization problem is convex and nonsmooth since the $\ell_{1,\infty}$ norm is nondifferentiable. Here there is a global minimum at $\bbeta$ if and only if the subdifferential of the objective function at $\bbeta$ contains the $\textbf{0}$-vector. The optimality conditions for \eqref{grlasso2} are given by
\begin{eqnarray}
\Rb^{n-1} \bbeta - \br^n + \lambda^n \bz = \textbf{0}, \quad \bz \in \partial \|\bbeta \|_{1,\infty}. \label{optcond}
\end{eqnarray}
We also have that $\bz \in \partial \|\bbeta \|_{1,\infty}$ if and only if $\bz$ satisfies the following conditions,
\begin{eqnarray}
\|\bz_{\cA_j}\|_1 &=& 1, \quad j \in \cP, \label{sub1} \\
\mathrm{sgn} (\bz_{\cA_j}) &=& \mathrm{sgn} (\bbeta_{\cA_j}) , \quad j \in \cP ,\label{sub2}\\
\bz_{\cB} &=& \textbf{0}, \label{sub3}\\
\|\bz_{\cC_j}\|_1 & \leq & 1, \quad j \in \cQ,\nonumber
\end{eqnarray}
where $\cA,\cB,\cC,\cP$ and $\cQ$ are $\bbeta$-dependent sets defined above. For notational convenience we leave out the time variable $n$ in the set notation. As $\bbeta_{\cC}=\textbf{0}$, \eqref{optcond} implies that
\begin{eqnarray}
\Rb^{n-1}_{\cS} \bbeta_{\cS} - \br^{n}_{\cS} + \lambda^n \bz_{\cS} = \textbf{0},\label{optcond1}\\
\Rb^{n-1}_{\cC \cS} \bbeta_{\cS} - \br^{n}_{\cC}+\lambda^n \bz_{\cC} = \textbf{0}.\nonumber
\end{eqnarray}
If $\Rb^{n-1}_{\cS}$ is invertible, then the solution is unique and we can rewrite \eqref{optcond1} as
\begin{eqnarray}
\bbeta_{\cS} = (\Rb^{n-1}_{\cS})^{-1} (\br^{n}_{\cS} - \lambda^n \bz_{\cS}). \label{optcond2}
\end{eqnarray}
Let $\Xb^{n-1} \in \RR^{n \times m}$ be the design matrix at time $n$ defined as 
\begin{eqnarray}
(\Xb^{n-1})^T := [\bx^0,\bx^1,\cdots,\bx^{n-1}],\nonumber
\end{eqnarray} 
and 
\begin{eqnarray}
\Yb^n:= [y^1,\ldots,y^n]^T.\nonumber
\end{eqnarray}
Then \eqref{optcond2} is equivalent to 
\begin{eqnarray}
\bbeta_{\cS} = \left[ (\Xb^{n-1}_{\ast \cS})^T \Xb^{n-1}_{\ast \cS} \right]^{-1} \left[ (\Xb^{n-1}_{\ast\cS})^T \Yb^n - \lambda^n \bz_{\cS} \right]. \label{optcond3}
\end{eqnarray}
Let $\Mb_{\cS}^{n-1} =  \left[ (\Xb^{n-1}_{\ast \cS})^T \Xb^{n-1}_{\ast \cS} \right]^{-1}$. Since the elements of $\Yb^n$ are independent and $\mathrm{Cov}(\Yb^n) = \sigma_\epsilon^2 \Ib$, \eqref{optcond3} gives us
\begin{eqnarray}
\mathrm{Cov} (\bbeta_{\cS})^{(n)}=\Mb_{\cS}^{n-1} \sigma_\epsilon^2 + ( \lambda^n)^2 \Mb_{\cS}^{n-1} \mathrm{Cov} (\bz_{\cS})^{(n)} \Mb_{\cS}^{n-1}.\label{optcond4}
\end{eqnarray}
By definition, $\hat{\bSigma}^{\bvartheta,n}_{\cS} :=\mathrm{Cov} (\bbeta_{\cS})^{(n)}$. If we replace $n$ with $n+1$, \eqref{optcond4} provides us with the equation
\begin{eqnarray}
\hat{\bSigma}^{\bvartheta,n+1}_{\cS} = \Mb_{\cS}^{n} \sigma_\epsilon^2 + ( \lambda^{n+1})^2 \Mb_{\cS}^{n} \mathrm{Cov}(\bz_{\cS})^{(n+1)}  \Mb_{\cS}^{n}.\label{optcond5}
\end{eqnarray}
One should note that we can not directly compute $\hat{\bSigma}^{\bvartheta,n+1}_{\cS}$ from the right hand side of \eqref{optcond5}, since $\bz_{\cS}$ is also a random variable dependent on $\hat{\bvartheta}^{n+1}_{\cS}$. But assuming that $\hat{\bvartheta}^{n+1}_{\cS}$ should not be far from $\bvartheta^{n}_{\cS}$, one can sample a set of random variables from the distribution $\cN(\bvartheta^n_{\cS}, \bSigma^{\bvartheta, n}_{\cS})$ and then sample the subgradients according to the equations \eqref{sub1}, \eqref{sub2} and \eqref{sub3}, so $\mathrm{Cov} (\bz_{\cS})^{(n+1)}$ can be estimated from the sample covariance matrix estimator $\hat{\mathrm{Cov}} (\bz_{\cS})^{(n+1)}$. Additionally, to make this estimator stable in theory, we need to make sure that all the eigenvalues of $\hat{\mathrm{Cov}} (\bz_{\cS})^{(n+1)}$ are bounded away from 0 and infinity. Heuristically, we first define a matrix space $\cM(C_{\min}, C_{\max})$ as
\begin{eqnarray}
\cM(C_{\min}, C_{\max}) = \{\Mb: C_{\min} \leq \Lambda_{\min}(\Mb) \leq \Lambda_{\max}(\Mb) \leq C_{\max} \}.\nonumber
\end{eqnarray}
Then we can project $\hat{\mathrm{Cov}} (\bz_{\cS})^{(n+1)}$ into $\cM(C_{\min}, C_{\max})$ and find a solution $\tilde{\mathrm{Cov}} (\bz_{\cS})^{(n+1)}$ to the following convex optimization problem
\begin{eqnarray}
\tilde{\mathrm{Cov}} (\bz_{\cS})^{(n+1)} = \argmin_{\Mb \in \cM(C_{\min}, C_{\max})} \|\hat{\mathrm{Cov}} (\bz_{\cS})^{(n+1)} - \Mb\|_{F}.\label{covbound}
\end{eqnarray}
Empirically we can use a surrogate projection procedure that computes a singular value decomposition of $\hat{\mathrm{Cov}} (\bz_{\cS})^{(n+1)}$ and truncates all the eigenvalues to be within interval $[C_{\min}, C_{\max}]$. Therefore we can approximately estimate $\hat{\bSigma}^{\bvartheta,n+1}_{\cS}$ by
\begin{eqnarray}
\hat{\bSigma}^{\bvartheta,n+1}_{\cS} = \Mb_{\cS}^{n} \sigma_\epsilon^2 + ( \lambda^{n+1})^2 \Mb_{\cS}^{n} \tilde{\mathrm{Cov}} (\bz_{\cS})^{n+1} \Mb_{\cS}^{n}. \label{sigmaMC}
\end{eqnarray}

Now we have all the ingredients for the knowledge gradient policy for sparse linear model (KGSpLin) and we outline it in Algorithm \ref{Algorithm1}.

\noindent\begin{minipage*}{\textwidth}
\setcounter{mpfootnote}{\value{footnote}}
\renewcommand{\thempfootnote}{\arabic{mpfootnote}}
\begin{algorithm}[H]
\textbf{Input:} $\bvartheta^0, \bSigma^{\bvartheta,0}, \{\xi_j^0, \eta_j^0\}_{j=1}^{p}, \tilde{\Xb}, \{\lambda^i\}_{i=1}^N.$\\
\textbf{Output}: $\bvartheta^N, \bSigma^{\bvartheta,N}, \{\xi_j^N, \eta_j^N\}_{j=1}^{p}$.\\
\For {$n = 0:N-1$}{
1. KG: $\bx^n = \argmax v^{KG,n}_x$;\\
2. Lasso homotopy update:\footnote{In practice, we often begin with some historical observations. Thus in the first iteration the Lasso estimator can be obtained from the historical dataset.} $\hat{\bvartheta}^n, (\bx^n,y^{n+1}) \in \RR^m \times \RR, \lambda^n,\lambda^{n+1} \rightarrow \hat{\bvartheta}^{n+1}$; \\
3. Monte Carlo Simulation: approximately simulate $\hat{\bSigma}_{\cS}^{\bvartheta,n+1}$ from \eqref{sigmaMC};\\
4. Bayesian update to: $ \bvartheta^{n+1}, \bSigma^{\bvartheta,n+1}, \{\xi_j^{n+1}, \eta_j^{n+1}\}_{j=1}^{p}$.\\
}
\caption{Sparse Knowledge Gradient Algorithm}\label{Algorithm1}
\end{algorithm}
\end{minipage*}

\section{Knowledge Gradient for Sparse Additive Model}\label{KGSpAM}

As we have the sparse knowledge gradient algorithm for $\ell_{1,\infty}$ group Lasso, we can generalize the knowlege gradient for sparse linear model to a nonparametric sparse additive model. In this section, we first describe the knowledge gradient for a sparse additive model, then we generalize it to the multivariate functional ANOVA model through tensor product splines. 

\subsection{Sparse Additive Modeling}\label{1AM}
In the additive model, $\bmu=[\mu_1,\ldots,\mu_M]^T \in \RR^M$, $\tilde{\Xb} = [\tilde{X}_{ij}] \in \RR^{M \times p}$ is the alternative matrix and
\begin{eqnarray}
\mu_i = f(\tilde{\Xb}_{i\ast})=\varsigma_i+ \sum_{j=1}^p f_j(\tilde{X}_{ij}), \quad \text{for } i=1,\ldots,M, \label{am}
\end{eqnarray}
where the $f_j$s are one-dimensional smooth component functions, one for each covariate and $\bvarsigma = [\varsigma_1,\ldots,\varsigma_M]^T$ is the residual term. For simplicity and identification purposes, we assume $\bvarsigma = \textbf{0}$ and $\int f_j(x_j)\,\mathrm{d}x_j=0$ for each $j$. When $f_j(x) = \alpha_j x$, this simply reduces to the linear model in Section \ref{KGLin}. In a high dimensional setting, where $p$ may be relatively large, we assume most of the $f_j$s are zeros.

If the truth takes the nonparametric additive form as in \eqref{am}, similarly, we let the choice of which $f_j$ is selected or not be random. Let $\bzeta=[\zeta_1,\ldots,\zeta_p]^T \in \RR^p$ be the indicator random variable of $f_j$'s, that is,
\begin{eqnarray}
\zeta_j = \left\{
  \begin{array}{l l}
    1 & \quad \text{if $f_j \neq 0$}\\
    0 & \quad \text{if $f_j=0$}
  \end{array}, \right.
  \quad \text{for } j=1,\ldots,p.\nonumber
\end{eqnarray}

Firstly, let us approximate each functional component in \eqref{am} through one-dimensional splines. Without loss of generality, suppose that all elements of $\tilde{\Xb}$ take values in $[0,1]$.  Let $0=\tau_0<\tau_1<\cdots<\tau_K<\tau_{K+1}=1$ be a partition of $[0,1]$ into $K+1$ subintervals.  Let $\cS_l$ be the space of polynomial splines of order $l$ (or degree $l-1$) consisting of functions $h$ satisfying
\begin{itemize}
\item[1] the restriction of $h$ to each subinterval is a polynomial of degree $l-1$;
\item[2] for $l \geq 2$ and $0 \leq l' \leq l-2$, $h$ is $l'$ times continuously differentiable on $[0,1]$.
\end{itemize}
This definition is phrased after \citet*{stone1985}, which is a descriptive version of Definition 4.1 in \citet[p. 108]{schumaker1981spline}. Under suitable smoothness assumptions, the $f_j$'s can be well approximated by functions in $\cS_{l_j}$. Specifically, let $\tilde{f}_j \in \cS_{l_j} $ be the estimate of $f_j$. Furthermore, for each $\tilde{f}_j$, there exists a normalized B-spline basis $\{\phi_{jk}(x), 1 \leq k \leq d_{j} \}$ for $\cS_{l_j}$, where $d_j=K+l_j$ \citep{schumaker1981spline}. If we let $\balpha_{j\bullet} = [\alpha_{j1},\ldots,\alpha_{jd_j}]$ be the coefficients of $\tilde{f}_j$ projected onto $\cS_{l_j}$, then for any $\tilde{f}_j \in \cS_{l_j}$, we can write 
\begin{eqnarray}
\tilde{f}_j (x) = \sum_{k=1}^{d_j} \alpha_{jk} \phi_{jk}(x), \quad \text{for } 1 \leq j \leq p. \label{SPAMbasis}
\end{eqnarray}

\noindent\begin{minipage*}{\textwidth}
\setcounter{mpfootnote}{\value{footnote}}
\renewcommand{\thempfootnote}{\arabic{mpfootnote}}
\begin{algorithm}[H]
\caption{Knowledge Gradient Algorithm for Sparse Additive Models}\label{Algorithm2}
\textbf{Input:\footnote{The prior mean and covariance matrix can also be obtained by some priors on $f_j$'s.}} $\bvartheta^0, \bSigma^{\bvartheta,0}, \{\xi_j^0, \eta_j^0\}_{j=1}^p, \tilde{\Xb}, \{\lambda^i\}_{i=1}^N, \{\phi_{jk}\}_{k=1,j=1}^{d_j,p}, \{\tau_j\}_{j=0}^{K+1}$\\
\textbf{Output}: $\{f_j^N\}_{j=1}^p, \bvartheta^N, \bSigma^{\bvartheta,N}, \{\xi_j^N, \eta_j^N\}_{j=1}^p$.\\
\For {$n = 0:N-1$}{
1. KG: $\bx^n = \argmax v^{KG,n}_x$;\\
2. Lasso homotopy update: $\hat{\bvartheta}^n, (\phi_{jk}(x_j^n),y^{n+1}) \in \RR^m \times \RR, \lambda^n,\lambda^{n+1} \rightarrow \hat{\bvartheta}^{n+1}$; \\
3. Monte Carlo Simulation: approximately simulate $\hat{\bSigma}^{\bvartheta,n+1}$ from \eqref{sigmaMC};\\
4. Bayesian update to: $ \{f_j^{n+1}\}_{j=1}^p, \bvartheta^{n+1}, \bSigma^{\bvartheta,n+1}, \{\xi_j^{n+1}, \eta_j^{n+1}\}_{j=1}^p$.\\
}
\end{algorithm}
\end{minipage*}
\vspace{0.5cm}

Let $\balpha=[\balpha_{1\bullet},\ldots,\balpha_{p\bullet}]$. We assume that $\balpha$ takes the conditional distribution
\begin{eqnarray}
\balpha|\bzeta \sim \cN(\bvartheta, \bSigma^{\bvartheta}),\nonumber
\end{eqnarray}
and also has the sparsity structure as described in Section \ref{KGLin}. Then at time $n$, we also have estimate $\hat{f}_j^n$ from group Lasso based on one-dimensional splines. More Specifically, for each $\hat{f}_j^n \in \cS_{l_j}$, let $\hat{\bvartheta}^n_{j\bullet} = [\hat{\vartheta}^n_{j1},\ldots,\hat{\vartheta}^n_{jd_j} ]$ be the coefficients of $\hat{f}_j^n$ and let $\hat{\bvartheta}^n = [ \hat{\bvartheta}^n_{1\bullet},\ldots,\hat{\bvartheta}^n_{p\bullet}]$. Accordingly, in the batch setting, where we already have $n$ samples $(\bx^{i-1},y^i) \in \RR^m \times \RR, i=1,\ldots,n$, one can get $\hat{\bvartheta}^n$  by solving the following penalized least squares problem
\begin{eqnarray}
\hat{\bvartheta}^n = \argmin_{\bvartheta \in \RR^m} \frac{1}{2}\sum_{i=1}^n\left[y^i-\sum_{j=1}^p \sum_{k=1}^{d_j} \vartheta_{jk} \phi_{jk} (x^{i-1}_j) \right]^2+\lambda \sum_{j=1}^p \| \bvartheta_{j \bullet}\|_\infty, \label{amgrlasso}
\end{eqnarray}
where $\lambda$ is the tuning parameter. Optimization problem \eqref{amgrlasso} is essentially an $\ell_{1,\infty}$ group Lasso optimization problem. The parameter $p$ is the number of groups and the group sparse solution on $\hat{\bvartheta}$ would lead to a sparse solution on $f_j$'s. Accordingly, we can also derive the knowledge gradient policy and Bayesian updating formulas as in Section \ref{KGLin}. Here we let $f_j^n$ be the Bayesian estimate of $f_j$ at time $n$, that is,
\begin{eqnarray}
f_j^n (x) = \sum_{k=1}^{d_j} \vartheta_{jk}^n \phi_{jk}(x), \quad \text{for } 1 \leq j \leq p. \nonumber
\end{eqnarray}
We outline the knowledge gradient algorithm for sparse additive models (KGSpAM) in Algorithm \ref{Algorithm2}.

\subsection{Tensor Product Smoothing Splines Functional ANOVA}
If the regression functions in \eqref{am} can also take bivariate or even multivariate functions, this model is known as the smoothing spline analysis of variance (SS-ANOVA) model \citep{wahba1990spline,wahba1995smoothing,gu2002smoothing}. In SS-ANOVA, we write
\begin{eqnarray}
\mu_i = f(\tilde{\Xb}_{i\ast})=\varsigma_i+\sum_{j=1}^p f_j(\tilde{X}_{ij}) + \sum_{j<k} f_{jk} (\tilde{X}_{ij}, \tilde{X}_{ik}) + \cdots, \label{ssanova}
\end{eqnarray}
where $f_j$'s are the main effects components, $f_{jk}$'s are the two-factor interaction components, and so on. $\bvarsigma$ is the residual term. Similar as before, we assume $\bvarsigma =\textbf{0}$, $\int f_j(x_j)\,\mathrm{d}x_j=0$ for each $j$, $\iint f_{jk} (x_j,x_k)\,\mathrm{d}x_j\mathrm{d}x_k = 0$ for each $j,k$ and so on. This model is also called functional ANOVA. The sequence is usually truncated somewhere to enhance interpretability. This SS-ANOVA generalizes the popular additive model in Section \ref{1AM} and provides a general framework for nonparametric multivariate function estimation, thus has been widely studied in the past decades.

As we approximate each $f_j$ by $\cS_{l_j}$, under certain smoothness assumptions, $f_{jk}$ can be well approximated by the tensor product space $\cS_{l_j} \otimes \cS_{l_k}$ defined by
\begin{eqnarray}
\cS_{l_j} \otimes \cS_{l_k} :&=& \{h_j h_k: \text{for all } h_j \in \cS_{l_j}, h_k \in \cS_{l_k}\}\nonumber\\
&=& \{ \sum_{r=1}^{d_j} \sum_{q=1}^{d_k} c_{rq} \phi_{jr}\phi_{kq}: \text{for all } c_{rq} \in \RR \}.\nonumber
\end{eqnarray}
Let
\begin{eqnarray}
\phi_{jrkq} (x_j,x_k) := \phi_{jr} (x_j) \phi_{kq} (x_k), \quad \text{for } 1 \leq r \leq d_j, 1 \leq q \leq d_k,\nonumber
\end{eqnarray}
then these are the basis functions for $d_j d_k$ dimensional tensor product space $\cS_{l_j} \otimes \cS_{l_k}$. This can also be generalized to multi-factor interaction components. Therefore, similarly, we can write all the functional components in \eqref{ssanova} as basis expansion forms. Then we can generalize a knowledge gradient algorithm for SS-ANOVA model.

\section{Theoretical Results}\label{proof}

In this section we provide the estimation error bounds of the Bayesian posterior mean estimate in Algorithm \ref{Algorithm1} as well as of the functional estimate in Algorithm \ref{Algorithm2}. We first state the selection and estimation properties of $\ell_{1,\infty}$ group Lasso in high dimensional settings when the number of groups exceeds the sample size. We show the estimation error bound of group Lasso. We also provide the sufficient conditions under which the group Lasso selects a model whose dimension is comparable with the underlying model with high probability. Based on these results, we assume that we begin with some historical observations and the Lasso estimator from the historical dataset has good initial property. If we have such a ``warm" start, we can show that the Bayesian posterior estimation error is bounded as in Theorem \ref{thm:bound}. The theorem actually shows that the posterior can converge to the truth at the same rate as that of group Lasso. Besides, based on this error bound, we can also show the estimation error bound of the functional estimate as in Theorem \ref{thm:fbound}. Note that these error bounds are proved on the intersection $\bar{\cS}$ of the support set $\cS^n$ from group Lasso estimator. But we can also prove that $\bar{\cS}$ is comparable with the true support set $\cS^{\ast}$. Additionally, all these theorems show the estimation error bounds as large enough measurements are made. Since our policy is also myopically optimal by construction, this lends a strong theoretical guarantee that the algorithm will work well for finite budgets.

\subsection{Bayesian Posterior Mean Estimation Error Bound}\label{sec::proof1}
In addition to the aforementioned notation, let $\bepsilon^n =[\epsilon^1,\ldots,\epsilon^n]^T$ be the measurement noise vector, so we have $\Yb^n:=\Xb^{n-1} \bvartheta +\bepsilon^n$. Then, we define the maximum group size $\bar{d} := \max_{j=1,\ldots,p}  d_j$ and the minimum group size $\underline{d} :=\min_{j=1,\ldots,p}  d_j$. Let $d = \bar{d}/\underline{d}$. Let $\cS^n = \{j: \hat{\bvartheta}^n_{\cG_j} \neq 0 \}$ be the estimated group support from current Lasso estimator. Let $\cS^{\ast}$ be the true support. Also, let $s^{\ast}=|\cS^{\ast} |$ be the cardinality of $\cS^{\ast}$. 

Before proving the estimation error bound, let us first introduce the selection and estimation properties of $\ell_{1,\infty}$ group Lasso. Our presentation will need the following assumptions. 

\begin{assumption}\label{subgaussian}
For any $n$, the random noise errors $\epsilon^1,\ldots,\epsilon^n$ are independent and identically distributed as $\cN(0,\sigma_{\epsilon}^2)$.
\end{assumption}

\begin{assumption}\label{RE}
The design matrix $\Xb^{n-1}$ satisfies the sparse Riesz condition (SRC) with rank $r$ and spectrum bounds $0<c_{\ast} < c^{\ast} < \infty$ if
\begin{eqnarray}
c_{\ast} \|\bnu \|_2^2\leq \frac{\|\Xb_{\ast \cS}^{n-1}\bnu \|_2^2}{n} \leq c^{\ast} \|\bnu \|_2^2, \quad \forall \cS \text{ with } r=|\cS| \text{ and } \bnu \in \RR^{\sum_{j \in \cS} d_j}.\nonumber
\end{eqnarray}
We refer to this condition as SRC $(r, c_{\ast}, c^{\ast})$.
\end{assumption}

\begin{assumption}\label{normalization}
For a given group $\cG=\{\cG_1,\ldots,\cG_{p}\}$. We say $\Xb^{n-1}$ is block normalized if   
\begin{eqnarray}
\frac{\| \Xb_{\ast \cG_j}^{n-1}\|_{2}}{\sqrt{n}} \leq 1, \quad \text{for all } j=1,2,\ldots, p.\nonumber
\end{eqnarray}
\end{assumption}
\begin{remark}
In Assumption \ref{normalization}, we set the upper bound to one in order to simplify notation. This particular choice entails no loss of generality. Note that this assumption is a natural generalization of the column normalization condition. Specifically, if we have $m = p$ groups, each of size one, the matrix norm reduces to the vector norm on every column of $\Xb^{n-1}$.
\end{remark}

All three assumptions can be reasonably expected to hold in practice. Assumption \ref{subgaussian} is on the distribution of random noise.  The SRC in Assumption \ref{RE} assumes the eigenvalues of the sample covariance matrix $\bSigma_{\cS}^{\Xb,n-1}:=\frac{1}{n}(\Xb_{\ast \cS}^{n-1})^T \Xb_{\ast \cS}^{n-1}$ are bounded below from zero and above from infinity when the size of $\cS$ is no greater than $r$.  It is natural to ask whether such condition also holds for general matrices. In fact, \citet*{zhang2008sparsity} provides sufficient conditions for the sparse Riesz condition for both deterministic and random design matrices $\Xb$. As we consider the designs are deterministic in this work, we only present the sufficient condition for deterministic design matrices proved by \citet*{zhang2008sparsity} in the following proposition. 

\begin{proposition}
Suppose $\Xb^{n-1}$ is column standardized with $\|\Xb_{\ast j}^{n-1} \|_2^2/n=1$. Let $\rho_{jk} =(\Xb_{\ast j}^{n-1})^T \Xb_{\ast k}^{n-1}/n$ be the correlation. If
\begin{eqnarray}
\max_{|\cS|=r}\inf_{\kappa \geq 1}\left\{\sum_{j \in \cG_{\cS}}\left(\sum_{k \in \cG_{\cS}, k \neq j} |\rho_{jk}|^{\kappa/(\kappa-1)}  \right)^{\kappa-1} \right\}^{1/\kappa} \leq \delta <1, \nonumber
\end{eqnarray}
then the sparse Riesz condition in Assumption \ref{RE} holds with rank $r$ and spectrum bounds $c_{\ast} = 1-\delta$ and $c^{\ast} = 1+\delta$. In particular, Assumption \ref{RE} holds with $c_{\ast} = 1-\delta$ and $c^{\ast} = 1+\delta$ if
\begin{eqnarray}
\max_{1 \leq j <k \leq m} |\rho_{jk} | \leq \frac{\delta}{r-1}, \quad \delta <1.\nonumber
\end{eqnarray}
\end{proposition}

Based on these assumptions, we can combine the results in \citet*{wei2010consistent} and \citet{negahban2012unified} and get the estimation error bound for $\ell_{1,\infty}$ group Lasso estimator as given in Lemma \ref{lassobound}.

\begin{lemma}\label{lassobound}
Under Assumption \ref{subgaussian}, \ref{RE} and \ref{normalization}, if we solve the group Lasso given in \eqref{grlasso} with
\begin{eqnarray}
\lambda^n =O(\bar{d}\sqrt{n\log p}), \nonumber
\end{eqnarray} 
then the following properties hold with probability converging to 1:
\begin{itemize}
\item[(1)] $|\cS^n| \leq C_1 |\cS^{\ast}|$ for some finite positive constant $C_1$. In specific, $C_1 = 2+4d\bar{c}$, where $\bar{c}:=c^{\ast}/c_{\ast}$.
\item[(2)] Any optimal solution $\hat{\bbeta}^n$ to \eqref{grlasso} satisfies the following error bound
 \begin{eqnarray}
 \|\hat{\bbeta}^n-\bbeta \|_2^2 \leq  \frac{C_2\sigma_{\epsilon}^2  s^{\ast}\bar{d}^2 \log p}{n}\nonumber,
 \end{eqnarray}
 for some positive constant $C_2$.
\end{itemize}
\end{lemma}

As one can see from the updating equations in \eqref{meanupdate} and \eqref{sigmaupdate}, the posterior mean estimate $\bvartheta_\cS^{n+1}$ is the weighted sum of prior $\bvartheta_\cS^{n}$ and the current Lasso estimate $\hat{\bvartheta}_{\cS}^{n+1}$. If the Lasso estimate has $\ell_2$ estimation bound as described in Lemma \ref{lassobound}, the posterior estimate should also have a similar bound under certain conditions of the weighted covariance matrix. One should note that both the mean and covariance are updated on some support $\cS$ from the current Lasso estimate. Thus we will work on a sequence of Lasso solutions and prove the bound on the intersection support set as large enough samples are made. Also note that in order to use the bound in Lemma \ref{lassobound}, we need to make sure that assumptions \ref{subgaussian}, \ref{RE} and \ref{normalization} are satisfied for every Lasso problem in such a sequence. Assumptions \ref{subgaussian} and \ref{normalization} are easy to satisfy. To show all the sequential Lasso problems satisfy Assumption \ref{RE}, we work from a ``warm" start at time $N'$. The following proposition actually verifies that if the design matrix at time $N'$ satisfies Assumption \ref{RE}, then the following ones should also satisfy this assumption, only with a slight loose on the constant. 

\begin{proposition} \label{REprop}
If for any $n$, there exists some constant $B >0$ such that $\|\bx^n\|_2^2\leq B$. Besides, assume for some large enough $N'$, the design matrix $\Xb^{N'-1}$ satisfies condition SRC $(r, c_{\ast}, c^{\ast})$. Then, for all $N'<n' \leq cN'$, of which $c>1$ is some constant, the design matrix $\Xb^{n'-1}$  can satisfy condition SRC $(r, c_{\ast}/c, \max(c^{\ast},B))$.
\end{proposition}

Thus we have all the ingredients to complete the proof of $\ell_2$ error bound of the Bayesian posterior mean estimator. Before that, let us state some assumptions for this result.
\begin{assumption}\label{2normbound}
For any $n$, there exists some constant $B >0$ such that $\|\bx^n\|_2^2\leq B$.
\end{assumption}
\begin{assumption} \label{REinitial}
For some large enough $n$, suppose for some constant $c>1$ and $n \leq cN'$, the design matrix $\Xb^{N'-1}$ satisfies the block normalization condition \ref{normalization} and condition SRC $(C_3 s^{\ast}, c_{\ast}, c^{\ast})$, where $C_3 := 2+4dc\max(c^{\ast},B)/c_{\ast}$.
\end{assumption}

Under these assumptions, we have the following theorem of the $\ell_2$ mean posterior estimation error bound.
\begin{theorem} \label{thm:bound}
Under Assumption \ref{subgaussian}, \ref{2normbound} and \ref{REinitial}, if we solve the group Lasso given in \eqref{grlasso} with
\begin{eqnarray}
\lambda^n =O(\bar{d}\sqrt{n\log p})\nonumber
\end{eqnarray} 
and let $\bar{\cS} :=\bigcap_{n'=N'}^n \cS^{n'}$, then the following properties hold with probability converging to 1:
\begin{itemize}
\item[(1)] $|\bar{\cS}| \leq C_3 |\cS^{\ast}|$ for some finite positive constant $C_3$ defined in Assumption \ref{REinitial}.
\item[(2)] Any posterior estimate $\bvartheta^n$ from Algorithm \ref{Algorithm1} satisfies
 \begin{eqnarray}
 \| \bvartheta_{\bar{\cS}}^n - \bvartheta_{\bar{\cS}}\|_2^2 \leq \frac{C_4 \sigma_{\epsilon}^2 s^{\ast} \bar{d}^2\log p}{n},\nonumber
 \end{eqnarray}
 for some positive constant $C_4$.
\end{itemize}
\end{theorem}

\subsection{Functional Estimation Error Bound}\label{sec::proof2}
Based on the results in Section \ref{sec::proof1}, we can also get the error bound for functional estimate of Algorithm \ref{Algorithm2} in Section \ref{1AM}. To show this error bound, let us introduce more definitions and assumptions.

Let $\beta$ be a nonnegative integer, let $\delta \in [0,1]$ be such that $q=\beta + \delta > 0.5$, and $L \in (0,\infty)$. Let $\cH (q, L)$ denote the collection of functions $h$ on [0,1] whose $\beta$th derivative, $h^{(\beta)}$, exists and satisfies the H$\ddot{\text{o}}$lder condition with exponent $\delta$,
\begin{eqnarray}
|h^{(\beta)}(t') - h^{(\beta)} (t)| \leq L |t'-t |^{\delta}, \quad \text{for } 0\leq t, t' \leq 1.\nonumber
\end{eqnarray}
Whenever the integral exists, for a function $h$ on $[0,1]$, denote its $\|\cdot\|_2$ norm by
\begin{eqnarray}
\|h\|_{2} := \sqrt{\int_0^1 h^2(x)\mathrm{d}x},\nonumber
\end{eqnarray}
Additionally, for any $\cS \subset \{1,\ldots,p\}$, we define
\begin{eqnarray}
\|h_{\cS}\|_2^2 := \sum_{j \in \cS} \|h_j\|_{2}^2.\nonumber
\end{eqnarray}
To prove the functional estimation error bound, we assume the true functions belong to this function class with smoothness parameter $q=2$.
\begin{assumption}\label{holder}
$f_j \in \cH(2,L)$ for $1 \leq j \leq p.$
\end{assumption}

Also note here we have the new design matrix $\Xb^{n-1}$ on the basis $\phi_{jk}$. Let $\Psi_j^{n-1}$ be the $n \times d_j$ matrix $\Psi_j (i,k) = \psi_{jk} (x^{i-1}_j)$, where $\psi_{jk}$ is the orthonormal B-spline basis. Let $\Psi^{n-1} := [\Psi_1^{n-1}, \ldots, \Psi_p^{n-1}]$.
Based on this and Theorem \ref{thm:bound}, we have the following theorem of the functional estimation error bound. 
\begin{theorem}\label{thm:fbound}
Under assumptions \ref{subgaussian} and \ref{holder}, if the design matrix $\Psi^{N'-1}$ satisfies Assumption \ref{REinitial} and \ref{2normbound}, the group Lasso is solved with some $\lambda^{n}$ satisfying
\begin{eqnarray}
\lambda^n =O(\bar{d}\sqrt{n\log p}), \nonumber
\end{eqnarray} 
let $\bar{\cS}:=\bigcap_{n'=N'}^n \cS^{n'}$, $\bar{d} = O(n^{1/6})$, $s^{\ast} = O(1)$, then the following properties hold with probability converging to 1:
\begin{itemize}
\item[(1)]  $|\bar{\cS} | \leq C_3 |\cS^{\ast}|$ for some finite positive constant $C_3$.
\item[(2)] Any posterior estimate $f^n$ from Algorithm \ref{Algorithm2} satisfies
\begin{eqnarray}
\|f_{\bar{\cS} }^n - f_{\bar{\cS} } \|_2^2 \leq \frac{C_5\sigma_{\epsilon}^2 \log p}{n^{2/3}},\nonumber
\end{eqnarray}
where $C_5$ is some positive constant.
\end{itemize}
\end{theorem}

\begin{remark}
Note here we use $\ell_{1,\infty}$ group Lasso instead of $\ell_{1,2}$ group Lasso, this is because the homotopy algorithm for recursive $\ell_{1,\infty}$ group Lasso largely reduces the computational complexity, but we do not have such results for $\ell_{1,2}$ group Lasso. However for $\ell_{1,2}$ group Lasso, the bound takes the form $\|\hat{\bbeta}^n-\bbeta \|_2^2 \precsim \frac{s^{\ast}\bar{d} \log p}{n}$. As one can see, the error term for $\ell_{1,\infty}$ group Lasso $\frac{s^{\ast}\bar{d}^2\log p}{n}$ is larger by a factor of $\bar{d}$, which corresponds to the amount by which an $\ell_{\infty}$-ball in $\bar{d}$ dimensions is larger than the corresponding $\ell_2$-ball. Therefore, we do not achieve the minimax optimal rate as in $\ell_{1,2}$ group Lasso. Thus using $\ell_{1,\infty}$ group Lasso instead of $\ell_{1,2}$ group Lasso is actually a tradeoff between computational complexity and statistical estimation. 
\end{remark}

\section{Experimental Testing}\label{simulation}
In this section, we investigate the performance of KGSpLin and KGSpAM in controlled experiments. In these experiments, we repeatedly sample the truth from some distribution and compare different policies to see how well we are discovering the truth. 
 
We first test the KGSpLin by generating a linear model with $p=100$ predictors, in ten groups of ten. The last 80 predictors all have coefficients of zero. The coefficients of the first 2 groups, that is 20 predictors, are randomly sampled from a normal distribution with means from 11 to 30 respectively, with standard deviation of 30\% of the mean. We randomly choose $M=100$ alternatives from some Gaussian distribution. Finally, normal measurement noise with standard deviation $\epsilon$ is added to each observation. In our first experiment, we focus on the comparison with KGLin and exploration policies using a relatively large measurement budget $N=200$. 

Furthermore, of all the experiments in this paper, to make a fair comparison of KG and exploration, the updating scheme when using the exploration policy is as described in Section \ref{sec::update}. The only difference is that at each iteration, exploration randomly measures each alternative with the same probability, while KG chooses the one with maximum KG value. Also, we assume that we do not have any prior information on the sparsity structures. That is, $\xi_j^0 = \eta_j^0 = 1$, for $j = 1,\ldots,p$.

\begin{figure*}[!htb]
\centering
\begin{tabular}{c}
\includegraphics[scale=0.42]{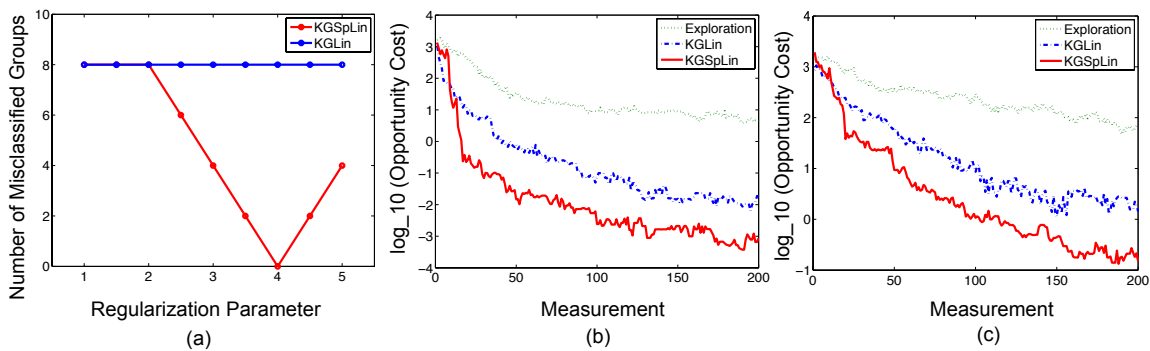}
\end{tabular}
\caption{(a) shows the misclassified groups for KGSpLin and KGLin as the regularization parameter $\lambda$ is varied. (b)(c) shows the averaged opportunity cost over 300 runs under low measurement noise (5\% range of the truth) and high measurement noise (30 \% range of the truth).}
\label{fig:1}
\end{figure*}

Figure \ref{fig:1}(a) shows the corresponding misclassification groups for KGSpLin and KGLin as the regularization parameter $\lambda$ is varied. (A misclassified group is one with at least one nonzero coefficient whose estimated coefficients are all set to zero, or vice versa.) Figure \ref{fig:1}(b) and (c) show the log of the averaged opportunity cost over 300 replications using a well chosen tuning parameter with low and high measurement noise (the standard deviations of the measurement noises are respectively 5\% and 30\% of the expected range of the truth). Here the opportunity cost (OC) is defined as the difference in true value between the best option and the option chosen by the policy, that is
\begin{eqnarray}
\mathrm{OC} = \max_i \mu_i - \mu_{i*}.\nonumber
\end{eqnarray}

From Figure \ref{fig:1} we can see that during the first several iterations, KGSpLin behaves comparable with pure exploration, because Lasso takes several iterations to identify the key features. However, after several initial samples, KGSpLin far outperforms both KGLin and pure exploration. This is because Lasso gives a rather precise estimate of the sparse linear coefficients given enough samples. So the algorithm mainly updates the beliefs on the key features based on these Lasso estimators, leading to more precise estimates of the model.  

To further compare the KGSpLin policy with KGLin from \citet{negoescu2011knowledge} for high dimensional sparse belief functions, we take several standard low dimensional test functions and hide them in a $p=200$ dimensional space. These functions were designed to be minimized, so both policies were applied to the negative of the functions. Table \ref{table2} shows the performance on the different functions.  Each policy was run 500 times with the specified amount of observation noise. Table \ref{table2} gives the sample mean and standard deviation of the mean of the opportunity cost after $N=50$ iterations of each policy. Here each function is scaled to have a range of 100, so that the measurement noises are given on the same scale.

\begin{table}[htbp]
\begin{center}\footnotesize
\renewcommand{\arraystretch}{1.3}
\begin{tabular}{cccccccc}
\hline
&&\multicolumn{3}{c}{{\bf KGSpLin}} & \multicolumn{3}{c}{{\bf
KGLin}} \\
Test function & $\sigma$&  $\mathbb{E}$(OC) & $\sigma$(OC) &  Med &  $\mathbb{E}$(OC) & $\sigma$(OC) & Med\\
\hline
Matyas                   &1 & .0104  &.0256 & .0071  & .0284 & .0157 & .0244\\
$\cX=[-10,10]^2$ & 10 &  .2772  & .1960 & .0125  & .3451 & .1166 & 0.3781\\
                              & 20 & .7658  & .8423 & .3997 & 1.7155 & .3208 & 1.5627\\
                              \hline
Trid                                &1 &   2.1422& 1.4011& 1.1843  & 2.7092 & 1.5331  & 1.3036\\
$d=6, \cX=[-36,36]^6$ & 10 &  9.8196  &3.8757 & 8.9874  & 9.9787 & 4.2098 & 8.2282\\
                                      & 20 &  15.7164  &4.0201 & 14.9040  & 16.8911 &  4.5881 &  15.4959\\
                                      \hline
Bohachevsky                  &1 & .0746 & .0249& .0035  &  .0853 & .0370& .0013\\
$\cX=[-100,100]^2$    & 10 &  .3585 & 2.5349& .2876  &  .5611 & 2.7056& .2993\\
                                     & 20 & 1.8224  &3.230 & 1.5578  & 1.9668 & 3.696 & 1.7008\\
                                     \hline
Six-hump Camel                  &1 &  .0023 & .0019 & .0000  & .0117 & .8097 & .0000\\
$\cX=[-3,3] \times [-2,2]$ & 10 &  .0895 & .6332 & .0000  & .1293 & .6098 & .0000\\
                                           & 20 & .4922  &.2159 & .0215  & .6183 & .2696 & 0.0306\\                             
\hline \hline
\end{tabular}
\end{center}
\caption{Quantitative comparison for KGSpLin and KGLin on standard test functions. Each row summarizes 500 runs of each policy on the specified test function. We compute the mean, standard deviation and median of OC. Each function is scaled to have a range of 100 and the results are given for different levels of noise standard deviation.}\label{table2}
\end{table}

\begin{figure*}[!htb]
\centering
\begin{tabular}{c}
\includegraphics[scale=0.48]{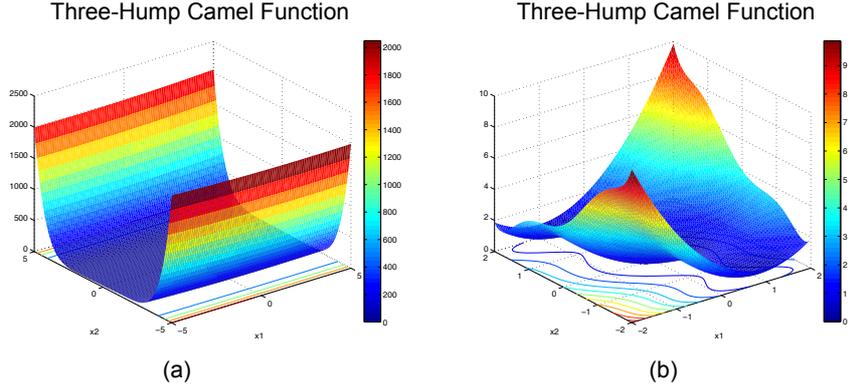}
\end{tabular}
\caption{(a) shows the negative Three-hump camel function on its recommended input domain, and (b) shows only a portion of this domain, to allow for easier viewing of the function's key characteristics. The function has one global maximum and two other local maxima.}
\label{fig:2}
\end{figure*}

Furthermore, we now test the KGSpAM policy on the following SS-ANOVA model with $p=100$ and four relevant variables,
\begin{eqnarray}
\mu_i = f_{12} (X_{i1}, X_{i2}) + \sum_{j=3}^5 f_j(X_{ij}) + \epsilon_i, \quad \epsilon_i \sim \cN(0, 1);\nonumber
\end{eqnarray}
the relevant component functions are given by
\begin{eqnarray}
f_{12} (x_1, x_2) &=& 2x_1^2 - 1.05 x_1^4 + \frac{x_1^6}{6} +x_1x_2+x_2^2,\label{spam12}\\
f_3(x)                 &=& 2 \sin(2 \pi x),\label{spam1}\\
f_4(x)                 &=& 8(x-0.5)^2,\label{spam2}\\
f_5(x)                 &=& 2\exp(-3x),\label{spam3}
\end{eqnarray}
where the first component function $f_{12}$ in \eqref{spam12} is known as the Three-hump camel function. We plot the true Three-hump camel function in Figure \ref{fig:2}(a), while the key part is shown in Figure \ref{fig:2}(b). For $f_{12}$, we use B-splines tensor product space $\cS_4 \otimes \cS_4$ to approximate it. The knot sequence is equally spaced on $[-5,5]^2$ with $K=4$ (the number of subintervals for each dimension is $K+1=5$). The remaining three relevant components are approximated using B-splines with order $l=4$ and equally spaced knot sequence on $[0,1]$ with $K=4$. The alternatives are uniformly sampled on the domain with $M=400$ and the measurement budget $N$ is 30. The standard deviation of measurement noise is set to 20\% of the expected range of the truth.

\begin{figure*}[!htb]
\centering
\begin{tabular}{c}
\includegraphics[scale=0.42]{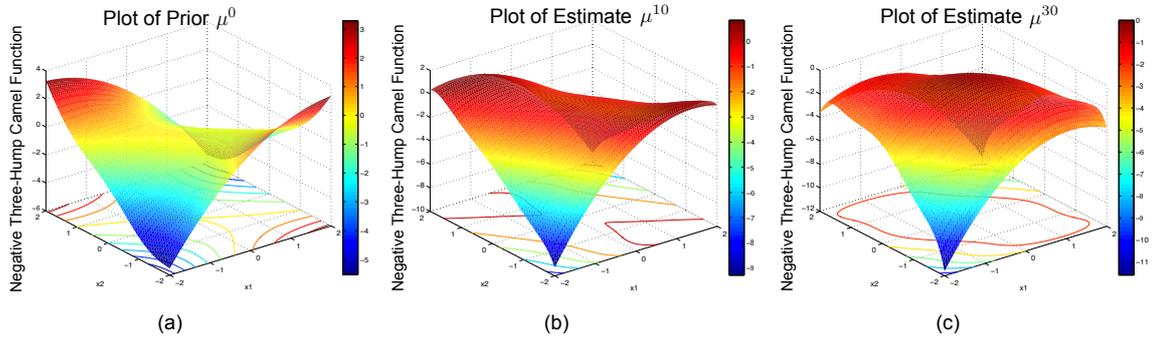}
\end{tabular}
\caption{(a) shows the prior of negative Three-hump camel function on its key region. (b) and (c) show the estimates of negative Three-hump camel function on its key region after 10 and 30 observations respectively.}
\label{fig:3}
\end{figure*}

\begin{figure*}[!htb]
\centering
\begin{tabular}{c}
\includegraphics[scale=0.42]{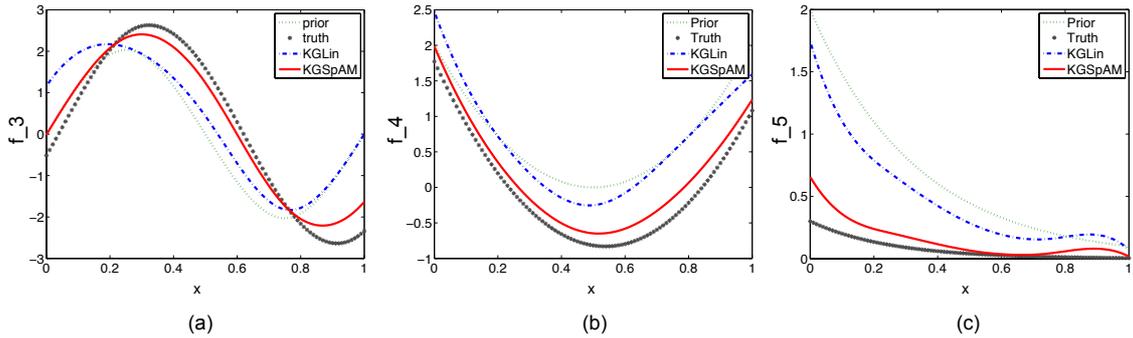}
\end{tabular}
\caption{(a)(b)(c) The prior, truth and final estimate of sparse additive model in \eqref{spam1}-\eqref{spam3} comparing KGLin and KGSpAM after $N=30$ observations. The standard deviation of measurement noise is 1, which is about 20\% of the expected range of the truth.}
\label{fig:4}
\end{figure*}

Then we run the KGSpAM policy on a $p=100$-dimensional space. To better visualize its performance, we plot the starting prior and estimated function of negative $f_{12}$ on its key region after the initial 10 and 30 observations as shown in Figure \ref{fig:3}. Comparing these estimates with the true function shown in Figure \ref{fig:2}, we visually see that the policy has done a good job estimating the lower key regions of the functions as desired after 10 observations and it identifies the areas of the three maxima after 30 observations. For the remaining three relevant functional components in \eqref{spam1}, \eqref{spam2} and \eqref{spam3}, we plot the prior, truth and final estimates of KGLin and KGSpAM in Figure \ref{fig:4}.

\section{Conclusion} In this paper, we extend the KG policy to high dimensional linear and nonparametric additive beliefs. It is a novel hybrid of Bayesian R\&S with the frequentist learning approach. Parallel with the Bayesian model, the policies use frequentist recursive Lasso approach to generate estimates and update the Bayesian model. Empirically, both KGSpLin and KGSpAM greatly reduce the measurement budget effort and  perform significantly better than several other policies in high dimensional setting. In addition, these policies are easy to implement and fast to compute. Theoretically, we prove that our policies are consistent. That is, the estimates can converge to the truth when given enough measurements. This also guarantees the convergence to global optimal alternative. All these advantages make them reasonable alternatives to other policies for high dimensional applications with sparse structure. Despite the advances, the convergence theory requires a number of structural assumptions, suggesting that future research should look to identify algorithms that work with more general model structures in high dimensions.

\appendix
\section*{Appendix A.}
\label{apx1}
Refer to Table \ref{table1}.
\begin{table}[htbp]
\begin{center}\footnotesize
\renewcommand{\arraystretch}{1.3}
\begin{tabular}{ll}\hline
\multicolumn{1}{l}{{\bf Variable}} & \multicolumn{1}{l}{{\bf Description}}\\ \hline
$\cX$ & Set of alternatives\\
$M$ & Number of alternatives\\
$N$ & Number of measurements budget\\
$\mu_x$ & Unknown mean of alternative $x$\\
$\sigma_x$ & Known standard deviation of alternative $x$\\
$\bmu$ & Column vector $(\mu_1,\ldots,\mu_M)^T$\\
$\bx^i$/$x^i$ & Sampling decision at time $i$ (vector or scalar index)\\
$\epsilon_{x}^{n+1}$ & Measurement error of alternative $\bx^n$\\
$y^{n+1}$ & Sampling observation from measuring alternative $\bx^n$\\
$\btheta^n$, $\bSigma^n$ & Mean and Covariance of prior distribution on $\mu$ at time $n$\\
$S^n$ & State variable, defined as the pair $(\btheta^n,\bSigma^n)$\\
$v_x^{KG,n}$ & Knowledge gradient value for alternative $x$ at time $n$\\
$\balpha$ & Vector of linear coefficients\\
$m$ & Number of features\\
$\tilde{\Xb}$ & Alternative matrix\\
$\bvartheta^n, \bSigma^{\bvartheta,n}$ & Mean of covariance of posterior distribution on $\alpha$ after $n$ measurements\\
$p$ & Number of nonoverlapping groups for features\\
$\cG, \cG_j$ & Group index\\
$d_j$ & Number of features in the $j$th group,$ d_j=|\cG_j|$\\
$\bzeta^n$ & Prior of $\bzeta$ at time $n$\\
$p_j^n$ & Parameter of Bernoulli distribution on $\zeta_j^n$\\
$(\xi_j^n, \eta_j^n)$ & Set of parameters of Beta distribution on $p_j^n$\\
$\hat{\bvartheta}^n$ & Lasso estimate at time $n$\\
$(\hat{\bvartheta}^{n}_{\cS}, \hat{\bSigma}^{\bvartheta,n}_{\cS})$ & Mean and covariance matrix estimator from Lasso solution at time $n$\\
$\cP^n$ & Index of selected groups from Lasso estimate at time $n$\\
$\cP$ & Active group index set\\
$\cQ$ & Inactive group index set\\
$\cA_j$ & Index set in the $j$th group with maximum absolute values\\
$\cB_j$ & Index set in the $j$th group except for $\cA_j$\\
$f_j$ & Smooth function of the $j$th feature\\
$K$ & Number of interior knots for one dimensional splines\\
$\cS_{l_j}$ & Space of polynomial spline of order $l_j$\\
$\phi_{jk}$ & $k$-th B-spline basis function for $\cS_{l_j}$ \\
$\alpha_{jk}$ & Coefficient for $f_j$ on basis function $\phi_{jk}$\\
$f_{jk}$ & Two-factor interaction component in SS-ANOVA model\\
$\phi_{jrkq}$ &  $rq$-th B-spline basis function for $\cS_{l_j} \otimes \cS_{l_k}$\\
$\bar{d}$ & Maximum group size\\
$\Xb^{n-1}$& Design matrix with rows of $\bx^0,\ldots,\bx^{n-1}$\\ 
$q$ & Smoothness parameter of the H$\ddot{\text{o}}$lder class $\cH$\\
$s^{\ast}$ & Cardinality of the true group set, $s^{\ast} = |\cS^{\ast}|$\\
\hline
\end{tabular}
\end{center}
\caption{Table of Notation}\label{table1}
\end{table}

\section*{Appendix B. Proofs}
In the following, we present the detailed proofs of all the technical results.
\subsection*{B.1 Proof of Proposition \ref{REprop}}
Let us define $\bSigma^{\Xb,n-1}$ be the sample covariance matrix, that is $\bSigma^{\Xb,n-1} = \frac{(\Xb^{n-1})^T \Xb^{n-1} }{n}$. 
For any $N'<n' \leq cN'$, let us divide the design matrix $\Xb^{n'-1}$,
\begin{eqnarray}
\Xb^{n'-1} = \begin{bmatrix}
       \Xb^{N'-1}\\
       \Xb^{+}     
     \end{bmatrix}.\nonumber
\end{eqnarray}
We need to prove $\Xb^{n'-1}$ satisfies condition SRC $(r, c_{\ast}/c, \max(c^{\ast},B))$. Note that $\Xb^{N'-1}$ satisfies SRC $(r, c_{\ast}, c^{\ast})$ is equivalent to
 \begin{eqnarray}
c_{\ast} \leq  \Lambda_{\min}(\Sigma^{\Xb,N'-1}_{\cS}) \leq \Lambda_{\max}(\Sigma^{\Xb,N'-1}_{\cS}) \leq c^{\ast}, \quad \forall \cS \text{ with } r=|\cS| \text{ and } \bnu \in \RR^{\sum_{j \in \cS} d_j}.\nonumber
\end{eqnarray}
Then we have that for $\forall \cS$ with $r=|\cS|$ 
\begin{eqnarray}
\bSigma_{\cS}^{\Xb, n'-1} &=& \frac{(\Xb_{\ast\cS}^{n'-1})^T \Xb_{\ast\cS}^{n'-1} }{n'} = \frac{(\Xb_{\ast\cS}^{N'-1})^T \Xb_{\ast\cS}^{N'-1} + (\Xb_{\ast\cS}^+)^T \Xb_{\ast\cS}^+ }{n'} \nonumber\\
&=&\frac{ N' \bSigma_{\cS}^{\Xb,N'-1} +  (\Xb_{\ast\cS}^+)^T \Xb_{\ast\cS}^+}{n'}\nonumber
\end{eqnarray}
This implies that
\begin{eqnarray}
\Lambda_{\min} (\Sigma^{\Xb,n'-1}_{\cS}) \geq \frac{N'}{n'} \Lambda_{\min}(\Sigma^{\Xb,N'-1}_{\cS}) \geq \frac{c_{\ast}}{c}\label{minbound}
\end{eqnarray}
and
\begin{eqnarray}
\Lambda_{\max} (\Sigma^{\Xb,n'-1}_{\cS}) \leq \frac{N'}{n'} \Lambda_{\max}(\Sigma^{\Xb,N'-1}_{\cS}) +\frac{1}{n'} \Lambda_{\max} [(\Xb_{\ast\cS}^+)^T \Xb_{\ast\cS}^+].\nonumber
\end{eqnarray}
Since 
\begin{eqnarray}
(\Xb_{\ast\cS}^+)^T \Xb_{\ast\cS}^+ = \bx_{\cS}^{N'} (\bx_{\cS}^{N'})^T + \bx_{\cS}^{N'+1} (\bx_{\cS}^{N'+1})^T+\cdots+\bx_{\cS}^{n'-1} (\bx_{\cS}^{n'-1})^T\nonumber
\end{eqnarray}
and
\begin{eqnarray}
\Lambda_{\max} [\bx_{\cS}^{n} (\bx_{\cS}^{n})^T ] = \|\bx_{\cS}^n\|_2^2 \leq B, \quad \forall n,\nonumber
\end{eqnarray}
we can get that
\begin{eqnarray}
\Lambda_{\max} (\Sigma^{\Xb,n'-1}_{\cS}) \leq \frac{N'}{n'} c^{\ast} + \frac{n'-N'}{n'} B \leq \max(c^{\ast},B).\label{maxbound}
\end{eqnarray}
Combining \eqref{minbound} and \eqref{maxbound} completes the proof. \endproof

\subsection*{B.2 Proof of Theorem \ref{thm:bound}}
The proof of part (1) directly follows Assumption \ref{REinitial}, Proposition \ref{REprop} and Lemma \ref{lassobound}. We now proceed to prove part (2).
If we let $\bar{\cS}:=\bigcap_{n'=N'}^n \cS^{n'}$, then from updating formula in \eqref{meanupdate} and \eqref{sigmaupdate}, we have
\begin{eqnarray}
\bvartheta^{n}_{\bar{\cS}} &=& \bSigma^{\bvartheta,n}_{\bar{\cS}} \left[(\bSigma^{\bvartheta,N'-1}_{\bar{\cS}})^{-1} \bvartheta^{N'-1}_{\bar{\cS}}+[(\hat{\bSigma}^{\bvartheta,N'}_{\cS^{N'}})^{-1}]_{\bar{\cS}} \hat{\bvartheta}^{N'}_{\bar{\cS}} + \cdots + [(\hat{\bSigma}^{\bvartheta,n}_{\cS^n})^{-1}]_{\bar{\cS}} \hat{\bvartheta}^{n}_{\bar{\cS}}  \right],\nonumber\\
\bSigma^{\bvartheta,n}_{\bar{\cS}} &=& \left[(\bSigma^{\bvartheta,N'-1}_{\bar{\cS}})^{-1} +[(\hat{\bSigma}^{\bvartheta,N'}_{\cS^{N'}})^{-1}]_{\bar{\cS}} + \cdots + [(\hat{\bSigma}^{\bvartheta,n}_{\cS^n})^{-1}]_{\bar{\cS}} \right]^{-1}.\nonumber
\end{eqnarray}
Then if we define
\begin{eqnarray}
\bdelta^{n'}_{\bar{\cS}} &:=& \bvartheta^{n'}_{\bar{\cS}} -\bvartheta_{\bar{\cS}} \nonumber\\
\hat{\bdelta}^{n'}_{\bar{\cS}} &:=& \hat{\bvartheta}^{n'}_{\bar{\cS}} -\bvartheta_{\bar{\cS}},\nonumber
\end{eqnarray}
for all $N'-1 \leq n' \leq n$ to simplify notation, we have
\begin{eqnarray}
\bdelta^{n}_{\bar{\cS}} = \bSigma^{\bvartheta,n}_{\bar{\cS}} \left[(\bSigma^{\bvartheta,N'-1}_{\bar{\cS}})^{-1} \bdelta^{N'-1}_{\bar{\cS}}+[(\hat{\bSigma}^{\bvartheta,N'}_{\cS^{N'}})^{-1}]_{\bar{\cS}}  \hat{\bdelta}^{N'}_{\bar{\cS}}+ \cdots + [(\hat{\bSigma}^{\bvartheta,n}_{\cS^n})^{-1}]_{\bar{\cS}}\hat{\bdelta}^{n}_{\bar{\cS}} \right].\nonumber
\end{eqnarray}
This gives us the following bound on $\bdelta^{n}_{\bar{\cS}}$,
\begin{multline}
\|\bdelta^{n}_{\bar{\cS}}\|_2 \leq \|\bSigma^{\bvartheta,n}_{\bar{\cS}}\|_2 \left[\|(\bSigma^{\bvartheta,N'-1}_{\bar{\cS}})^{-1}\|_2 \|\bdelta^{N'-1}_{\bar{\cS}}\|_2+\|[(\hat{\bSigma}^{\bvartheta,N'}_{\cS^{N'}})^{-1}]_{\bar{\cS}}\|_2 \|\hat{\bdelta}^{N'}_{\bar{\cS}}\|_2+ \right. \\ 
\cdots \left.+\|[(\hat{\bSigma}^{\bvartheta,n}_{\cS^n})^{-1}]_{\bar{\cS}}\|_2 \|\hat{\bdelta}^{n}_{\bar{\cS}}\|_2 \right].\nonumber
\end{multline}
We now proceed to bound each of the quantities. Let us for now assume that $N' \leq n' \leq n$.  As we suppose the design matrix for Lasso solution $\hat{\bvartheta}^{N'}_{\cS}$ satisfies Assumption \ref{REinitial}, by Proposition \ref{REprop} and Lemma \ref{lassobound}, if we choose $\lambda^{n'}$ such that
\begin{eqnarray}
\lambda^{n'} =O(\bar{d}\sqrt{n'\log p}),\label{eq:lambdaorder}
\end{eqnarray}
then there exists some constant $C_{6}$ such that
\begin{eqnarray}
\|\hat{\bdelta}^{n'}_{\bar{\cS}}\|_2 \leq C_{6} \sigma_\epsilon \bar{d} \sqrt{\frac{s^{\ast}\log p}{n'}}, \quad \text{for all }  N' \leq n' \leq n,\label{eq:bound1}
\end{eqnarray}
with probability converging to 1. We know from \eqref{sigmaMC} that
\begin{eqnarray}
\hat{\bSigma}^{\bvartheta,n'}_{\cS^{n'}} = \Mb_{\cS^{n'}}^{n'-1} \sigma_\epsilon^2 + ( \lambda^{n'})^2 \Mb_{\cS^{n'}}^{n'-1} \tilde{\mathrm{Cov}} {(\bz_{\cS^{n'}})}^{(n')} \Mb_{\cS^{n'}}^{n'-1},\nonumber
\end{eqnarray}
where
\begin{eqnarray}
\Mb_{\cS^{n'}}^{n'-1} =  \left[ (\Xb^{n'-1}_{\ast \cS^{n'}})^T \Xb^{n'-1}_{\ast \cS^{n'}} \right]^{-1}.\nonumber
\end{eqnarray}
Assumption \ref{REinitial} gives us
\begin{eqnarray}
\Lambda_{\max} (\Mb_{\cS}^{N'-1}) &\leq& \frac{1}{N' c_{\ast}} < \infty, \nonumber\\
\Lambda_{\min} (\Mb_{\cS}^{N'-1})  &\geq&  \frac{1}{N' c^{\ast}} > 0,\nonumber
\end{eqnarray} 
for any $\cS$ with $| \cS| = C_3 s^{\ast}$. Therefore, since $|\cS^{n'}| \leq C_3 s^{\ast}$, by Proposition \ref{REprop}, we can show that for all $N' \leq n' \leq n$, there exist positive constants $C_7$ and $C_8$, such that
\begin{eqnarray}
\Lambda_{\max} (\Mb_{\cS^{n'}}^{n'-1}) &\leq& \frac{C_7}{n'} < \infty, \label{eq:Mmax}\\
\Lambda_{\min} (\Mb_{\cS^{n'}}^{n'-1})  &\geq&  \frac{C_{8}}{n'} > 0. \label{eq:Mmin}
\end{eqnarray}
It is not hard to prove
\begin{eqnarray}
\Lambda_{\min} (\Mb \Nb) \geq \Lambda_{\min} (\Mb)\Lambda_{\min} (\Nb)\nonumber
\end{eqnarray}
for any positive semidefinite matrices $\Mb$ and $\Nb$, so using Weyl's inequality in matrix theory, \eqref{covbound} and \eqref{eq:Mmin}, we have the following bound,
\begin{eqnarray}
\|[(\hat{\bSigma}^{\bvartheta,n'}_{\cS^{n'}})^{-1}]_{\bar{\cS}}\|_2 &\leq& \|(\hat{\bSigma}^{\bvartheta,n'}_{\cS^{n'}})^{-1}\|_2 = \Lambda_{\min}^{-1}(\hat{\bSigma}^{\bvartheta,n'}_{\cS^{n'}})\nonumber\\
&\leq& \frac{1}{\Lambda_{\min} (\sigma_\epsilon^2 \Mb_{\cS^{n'}}^{n'-1}) +( \lambda^{n'})^2 \Lambda_{\min} (\hat{\mathrm{Cov}} (\bz^{n'}_{\cS^{n'}})) \Lambda_{\min}^2(\Mb_{\cS^{n'}}^{n'-1})}\nonumber\\
 &\leq& \frac{C_{9} n' }{\sigma_{\epsilon}^2\bar{d}^2\log p}, \label{eq:bound2}
\end{eqnarray}
for some constant $C_{9}$. Similarly, by \eqref{eq:lambdaorder}, \eqref{eq:Mmax}, and \eqref{covbound}, we can also get
\begin{eqnarray}
\|\hat{\bSigma}^{\bvartheta,n'}_{\cS^{n'}}\|_2 &=& \Lambda_{\max} (\hat{\bSigma}^{\bvartheta,n'}_{\cS^{n'}})\nonumber\\
& \leq& \sigma_\epsilon^2\Lambda_{\max}(\Mb_{\cS^{n'}}^{n'-1}) + ( \lambda^{n'})^2 \Lambda_{\max} (\hat{\mathrm{Cov}} (\bz^{n'}_{\cS^{n'}})) \Lambda_{\max}^2(\Mb_{\cS^{n'}}^{n'-1})\nonumber\\
&\leq& C_{10} \frac{\sigma_{\epsilon}^2 \bar{d}^2 \log p}{n'},\nonumber
\end{eqnarray}
for some constant $C_{10}$. Thus, for the posterior covariance matrix, we have
\begin{eqnarray}
\|\bSigma^{\bvartheta,n}_{\bar{\cS}}\|_2  &=& \Lambda^{-1}_{\min}\left[(\bSigma^{\bvartheta,N'-1}_{\bar{\cS}})^{-1} +[(\hat{\bSigma}^{\bvartheta,N'}_{\cS^{N'}})^{-1}]_{\bar{\cS}} + \cdots + [(\hat{\bSigma}^{\bvartheta,n}_{\cS^n})^{-1}]_{\bar{\cS}}\right]\nonumber\\
&\leq& \frac{1}{\Lambda_{\min}\left[ [(\hat{\bSigma}^{\bvartheta,N'}_{\cS^{N'}})^{-1}]_{\bar{\cS}}\right] +\cdots \Lambda_{\min}\left[(\hat{\bSigma}^{\bvartheta,n}_{\cS^n})^{-1}\right]_{\bar{\cS}}}\nonumber\\
&=&  \frac{1}{\Lambda^{-1}_{\max}(\hat{\bSigma}^{\bvartheta,N'}_{\cS^{N'}}) +\cdots \Lambda^{-1}_{\max} (\hat{\bSigma}^{\bvartheta,n}_{\cS^n})}\nonumber\\
&\leq& \frac{2C_{10} \sigma_{\epsilon}^2 \bar{d}^2 \log p}{(N'+n)(n-N'+1)}\nonumber\\
&\leq& \frac{C_{11} \sigma_{\epsilon}^2 \bar{d}^2 \log p}{n^2}, \label{eq:bound3}
\end{eqnarray}
for some constant $C_{11}$. If we let
\begin{eqnarray}
\Delta_{\bar{\cS}} (N') = \|(\bSigma^{\bvartheta,N'-1}_{\bar{\cS}})^{-1}\|_2 \|\bdelta^{N'-1}_{\bar{\cS}}\|_2, \nonumber
\end{eqnarray}
then combining \eqref{eq:bound1},\eqref{eq:bound2} and \eqref{eq:bound3} gives us the following bound on $\bdelta^{n}_{\bar{\cS}}$
\begin{eqnarray}
\|\bdelta^{n}_{\bar{\cS}}\|_2 &\leq&  \frac{C_{11} \sigma_{\epsilon}^2 \bar{d}^2 \log p}{n^2}\left(\Delta_{\bar{\cS}}(N')+ \sum_{n'=N'}^n \frac{C_{6} C_{9}\sqrt{s^{\ast}n'}}{\sigma_{\epsilon}\bar{d}\sqrt{\log p}}\right)\nonumber\\
&\leq& \frac{C_{12} \sigma_{\epsilon}\bar{d} \sqrt{s^{\ast}\log p}}{\sqrt{n}} + \frac{C_{11} \sigma_{\epsilon}^2 \bar{d}^2 \log p \Delta_{\bar{\cS}}(N')}{n^2}, \label{sumrate}
\end{eqnarray}
for some constant $C_{12}$, which is equivalent to
\begin{eqnarray}
\| \bvartheta_{\bar{\cS}}^n - \bvartheta_{\bar{\cS}}\|_2^2 \leq \frac{C_4 \sigma_{\epsilon}^2 s^{\ast} \bar{d}^2\log p}{n} \nonumber
\end{eqnarray}
and thus completes the proof.\endproof

\subsection*{B.3 Proof of Theorem \ref{thm:fbound}}
By definition of $f_j$, $1\leq j \leq p$, part (1) follows from part (2) of Theorem \ref{thm:bound} directly. Now consider part (2). We denote $\tilde{f}_j^{\ast}$ as
\begin{eqnarray}
\tilde{f}_j^{\ast} (x) = \sum_{k=1}^{d_j} \vartheta_{jk} \psi_{jk}(x), \quad \text{for } 1 \leq j \leq p. \nonumber
\end{eqnarray}
We also have
\begin{eqnarray}
f_j^n (x) = \sum_{k=1}^{d_j} \vartheta^n_{jk} \psi_{jk}(x), \quad \text{for } 1 \leq j \leq p.\nonumber
\end{eqnarray}
Since $\psi_{jk}$ is the orthonormal basis, we have
\begin{eqnarray}
\|f_j^n - \tilde{f}_j^{\ast}\|_2^2 \leq \|\bvartheta_{j \ast}^n - \bvartheta_{j \ast} \|_2^2.\nonumber
\end{eqnarray}
Also by Assumption \ref{holder} and Lemma 8 in \citet*{stone1986dimensionality}, taking $q=2$, we have
\begin{eqnarray}
\|\tilde{f}_j^{\ast} -f_j\|^2 = O(d_j^{-2q}) =O(d_j^{-4}).\nonumber
\end{eqnarray}
Thus by the result of Theorem \ref{thm:bound}, we have
\begin{eqnarray}
\| f_{\bar{\cS}}^n - f_{\bar{\cS}}\|^2 \leq \frac{C_{4}\sigma_{\epsilon}^2 s^{\ast} \bar{d}^2\log p}{n}+ \frac{C_{13}}{\bar{d}^4}\nonumber.
\end{eqnarray}
Note that choosing $\bar{d} = O(n^{1/6})$ and $s^{\ast} = O(1)$ would not change the rate in equation $\eqref{sumrate}$, so we have the following bound
\begin{eqnarray}
\|f_{\bar{\cS}}^n - f_{\bar{\cS}} \|_2^2 \leq \frac{C_5\sigma_{\epsilon}^2 \log p}{n^{2/3}}.\nonumber
\end{eqnarray}
\endproof

\medskip

\bibliographystyle{plainnat}
\bibliography{li14a}

\end{document}